\documentclass{article}

\usepackage[preprint, nonatbib]{main}

\usepackage[utf8]{inputenc} 
\usepackage[T1]{fontenc}    
\usepackage{hyperref}       
\usepackage{url}            
\usepackage{booktabs}       
\usepackage{amsfonts}       
\usepackage{nicefrac}       
\usepackage{microtype}      
\usepackage{graphicx}
\usepackage{subfigure}
\usepackage{algorithm}
\usepackage{algorithmic}
\usepackage{amsmath}
\usepackage{amssymb}

\title{Learning to Solve Combinatorial Optimization Problems on Real-World Graphs in Linear Time}

\author{
  Iddo Drori\\
  Dept. of CS, School of ORIE\\
  Columbia,
  Cornell University\\
  \texttt{idrori@cs.columbia.edu} \\
  \And
  Anant Kharkar \\
  Dept. of Computer Science\\
  Columbia University\\
  \texttt{agk2151@columbia.edu} \\
  \And
  William R. Sickinger \\
  Dept. of Computer Science\\
  Columbia University\\
  \texttt{wrs2125@columbia.edu}\\
  \And
  Brandon Kates \\
  Dept. of Computer Science\\
  Cornell University \\
  \texttt{bjk224@cornell.edu}\\
  \And
  Qiang Ma \\
  Dept. of Computer Science\\
  Columbia University\\
  \texttt{ma.qiang@columbia.edu} \\
  \And
  Suwen Ge \\
  Dept. of Computer Science\\
  Columbia University\\
  \texttt{sg3635@columbia.edu} \\
  \And
  Eden Dolev \\
  Dept. of Computer Science\\
  Columbia University\\
  \texttt{ed2566@columbia.edu} \\
  \And
  Brenda Dietrich \\
  School of ORIE\\
  Cornell University \\
  \texttt{bld34@cornell.edu} \\
  \And
  David P. Williamson \\
  School of ORIE\\
  Cornell University \\
  \texttt{davidpwilliamson@cornell.edu}\\
  \And
  Madeleine Udell \\
  School of ORIE\\
  Cornell University \\
  \texttt{udell@cornell.edu}\\
}

\begin{document}

\maketitle

\begin{abstract}
Combinatorial optimization algorithms for graph problems are usually designed afresh for each new problem with careful attention by an expert to the problem structure. In this work, we develop a new framework to solve any combinatorial optimization problem over graphs that can be formulated as a single player game defined by states, actions, and rewards, including minimum spanning tree, shortest paths, traveling salesman problem, and vehicle routing problem, without expert knowledge. Our method trains a graph neural network using reinforcement learning on an unlabeled training set of graphs. The trained network then outputs approximate solutions to new graph instances in linear running time. In contrast, previous approximation algorithms or heuristics tailored to NP-hard problems on graphs generally have at least quadratic running time. We demonstrate the applicability of our approach on both polynomial and NP-hard problems with optimality gaps close to 1, and show that our method is able to generalize well: (i) from training on small graphs to testing on large graphs; (ii) from training on random graphs of one type to testing on random graphs of another type; and (iii) from training on random graphs to running on real world graphs.
\end{abstract}

\pagebreak

\section{Introduction}
\label{sec:introduction}

A core and important area in computer science and operations research is the domain of graph algorithms and combinatorial optimization. The literature is rich in both exact (slow) and heuristic (fast) algorithms \cite{golden1980approximate}; however, each algorithm is designed afresh for each new problem with careful attention by an expert to the problem structure. Approximation algorithms for NP-hard problems provide only worst-case guarantees \cite{williamson2011design}, and are not usually linear time, and hence not scalable. Our motivation is to learn new heuristic algorithms for these problems that require an evaluation oracle for the problem as input and returns a good solution in a pre-specified time budget. Concretely, we target combinatorial and graph problems in increasing order of complexity, from polynomial problems such as minimum spanning tree (MST), and shortest paths (SSP), to NP-hard problems such as the traveling salesman problem (TSP) and the vehicle routing problem (VRP).

The aptitude of deep learning systems for solving combinatorial optimization problems has been demonstrated across a wide range of applications in the past several years \cite{hanjun2017learning, bengio2018machine}. Two recent surveys of reinforcement learning methods \cite{mazyavkina2020reinforcement} and machine learning methods \cite{vesselinova2020learning} for combinatorial optimization over graphs with applications have become available during the time of this writing. The power of graph neural networks (GNNs) \cite{xu2018powerful} and the algorithmic alignment between GNNs and combinatorial algorithms has recently been studied \cite{xu2020reason}. 
GNNs trained using specific aggregation functions emulate specific algorithms: 
for example a GNN aligns well \cite{xu2020reason} with the Bellman-Ford algorithm for shortest paths. 

Our work is motivated by recent theoretical and empirical results in reinforcement learning and graph neural networks:
\begin{itemize}
    \item GNN training is equivalent to a dynamic programming (DP) algorithm \cite{xu2020reason}, hence GNNs by themselves can be used to mimic algorithms with polynomial time complexity.
    \item Reinforcement learning methods with GNNs can be used to find approximate 
     solutions to NP-hard combinatorial optimization problems \cite{hanjun2017learning,bengio2018machine,kool2019attention}.
\end{itemize}

Combinatorial optimization problems may be solved 
by exact methods, by approximation algorithms, or by heuristics. 
Machine learning approaches for combinatorial optimization have mainly used either supervised or reinforcement learning. 
Our approach is unsupervised and is based on reinforcement learning. We require neither output labels nor knowing the optimal solutions, and our method improves by self play. Reinforcement learning methods can be divided into model-free and model-based methods.
In turn, model-free methods can be divided into Q-Learning and policy optimization methods \cite{rltaxanomy2020openai}. 
Model-based methods have two flavors: 
methods in which the model is given, such as expert iteration \cite{anthony2017thinking} or AlphaZero \cite{silver2017mastering}, 
and methods that learn the model, such as World Models \cite{ha2018world} or MuZero \cite{schrittwieser2019mastering}. AlphaZero has been generalized to many games \cite{cazenave2020polygames},
both multi-player and single-player \cite{drori2018alphad3m}. 
This work views algorithms on graphs as single-player games and learns graph algorithms. 
The supplementary material includes a comprehensive list of supervised and reinforcement learning methods used for combinatorial optimization of NP-Hard problems and a classification of all previous work by problem, method, and type. 

Previous work using reinforcement learning and a GNN representation to solve 
combinatorial optimization problems targets
individual problems with algorithms tailored for each one. 
In contrast, this work provides a general framework for model-free reinforcement learning using a GNN representation that elegantly adapts to different problem classes by changing an objective or reward.

Our approach generalizes well from examples on small graphs, 
where even exhaustive search is easy, to larger graphs; 
and the architecture works equally well when trained on polynomial problems 
such as minimum spanning tree (MST) as on NP-hard problems such as the traveling salesman problem (TSP), 
though training time is significantly larger for hard problems. 
We explore the limits of these algorithms as well: 
for what kinds of problem classes, problem instances, and time budgets
do they outperform classical approximation algorithms?

Our key contributions and conclusions are given below:

\begin{table}[b]
\centering
\small
\begin{tabular}{l|c|c|c|c}
\textbf{Method} & \textbf{Runtime Complexity} & \textbf{Runtime (ms)} & \textbf{Speedup} & \textbf{Optimality Gap}\\
\hline
Gurobi (Exact) & NA & $3,220$ & $2,752.1$ & $1$\\
Concorde (Exact) & NA & $254.1$ & $217.2$ & $1$\\
\hline
Christofides & $O(n^{3})$ & $5,002$ & $4,275.2$ & $1.029$\\
LKH & $O(n^{2.2})$ & $2,879$ & $2460.7$ & $1$\\
2-opt & $O(n^{2})$ & $30.08$ & $25.7$ & $1.097$\\
Farthest & $O(n^{2})$ & $8.35$ & $7.1$ & $1.075$\\
Nearest & $O(n^{2})$ & $9.35$ & $8$ & $1.245$\\
\hline
S2V-DQN & $O(n^{2})$ & $61.72$ & $52.8$ & $1.084$\\
GPN & $O(n\log n)$ & $1.537$ & $1.3$ & $1.086$\\
\hline
Ours & $O(n)$ & $1.17$ & $1$ & $1.074$\\
\hline
\end{tabular}
\caption{TSP running time complexity, runtime, speedup factor of our method, and optimality gap (smaller is better) versus optimal, averaged over 1000 
uniform random graphs each with 100 nodes.} 
\label{tab:tsp-key}
\end{table}

\begin{enumerate}

\item \textbf{Linear running time complexity with optimality gaps close to $1$.} 
In general for all graph problems, by construction our approximation running time is linear $O(n + m)$ in the number of nodes $n$ and edges $m$, both in theory and in practice. For MST and SSP our running time is linear $O(m)$ in the number of edges. For TSP and VRP our running time is linear $O(n)$ in the number of nodes. Table \ref{tab:tsp-key} summarizes our key results for TSP. 
TSP approximation algorithms and heuristics have runtimes that 
grow at least quadratically in the graph size. 

On random Euclidean graphs with 100 nodes, 
our method is one to three order of magnitude faster
to deliver a comparable optimality gap,
and this speedup improves as the graph size increases.
S2V-DQN \cite{hanjun2017learning}, another reinforcement learning method, builds a 10-nearest neighbor graph, 
and also has quadratic runtime complexity;
on these graphs, our method runs $52$ times faster and obtains a lower (better) optimality gap, which is the ratio between a method's reward and the optimal reward. 
GPN \cite{ma2020combinatorial} has runtime complexity $O(n \log n)$ with a larger optimality gap and does not generalize as well nor easily extend to other problems. 

The running time for solving MST using Prim's algorithm is $O(m \log m)$ and the running time for solving SSP using Dijkstra's algorithm is $O(n \log n + m)$. For MST, running our method on larger graphs (for longer times) results in optimality gaps that are close to 1, converging to an optimal solution. 
Figures \ref{fig:mst-optimality-gap} shows our results and optimality gaps for MST.

\item \textbf{Generalization on graphs.} (i) From small to large random graphs: For MST, we generalize from small to large graphs accurately. For TSP, we generalize from small to larger graphs with median tour lengths (and optimality gaps) that are better than other methods; (ii) Between different types of random graphs: For MST, we generalize accurately between different types of random graphs as shown in Figure \ref{fig:graph-types}; and (iii) From random to real-world graphs: For TSP, we generalize from random graphs to real-world graphs better than other methods,
as shown in Table \ref{tab:tsp-real-world}.

\item \textbf{A unified framework for solving any combinatorial optimization problem over graphs.}
(A) We model problems that involve both actions on nodes and edges by using the edge-to-vertex \emph{line graph}. Figure \ref{fig:unified-framework}(A) shows an example of a primal graph and its line graph. (B) We model graph algorithms as a single player game as shown in Figure \ref{fig:unified-framework}(B). (C) We learn different problems by changing the objective or reward function as shown in Figure \ref{fig:unified-framework}(C).

\end{enumerate}

\section{Unified Framework}

\begin{figure}[b!]
\centering
\includegraphics[width=0.9\textwidth]{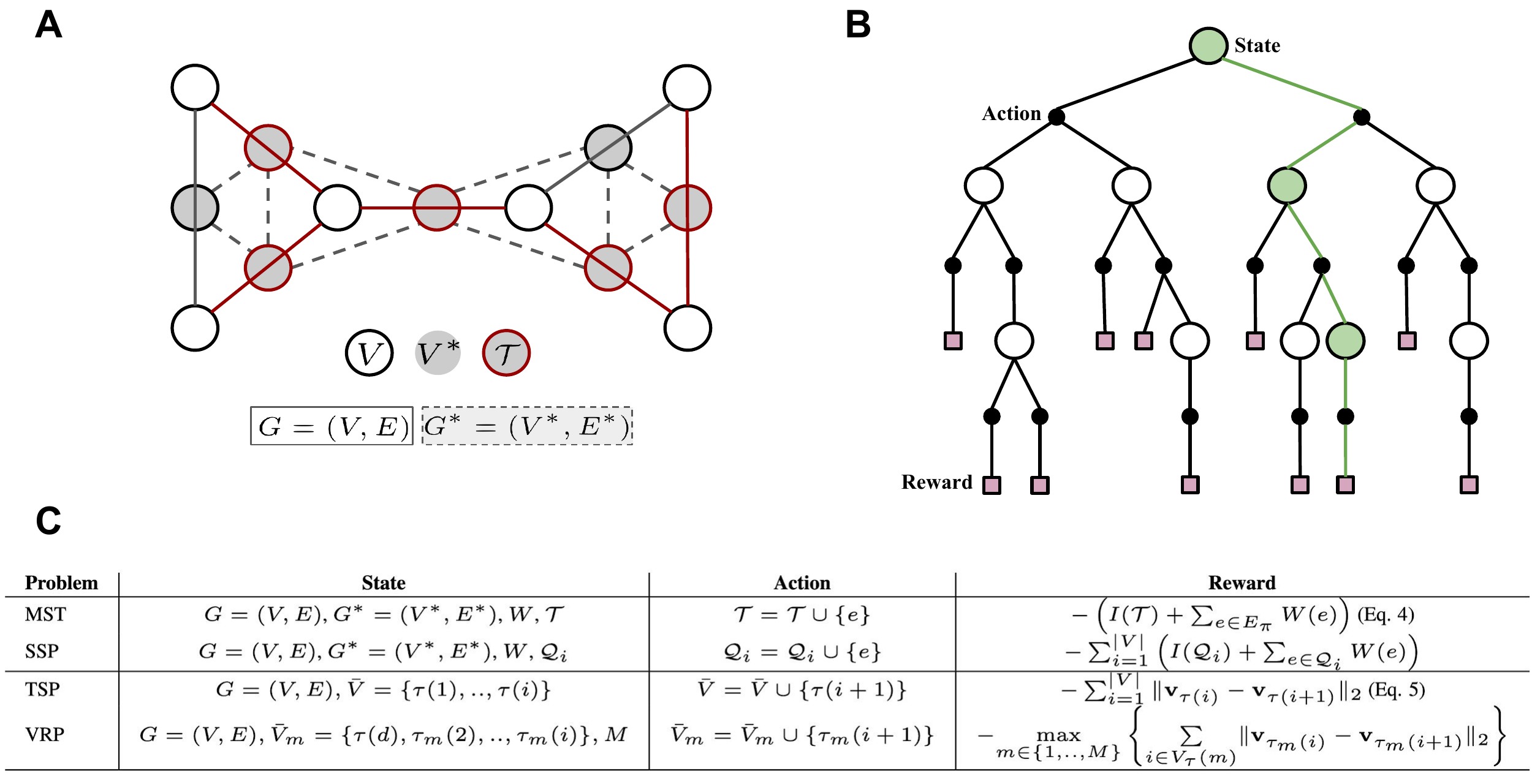}
\caption{Our unified framework: (A) The primal graph (white nodes and solid edges) and its edge-to-vertex line graph (gray nodes and dashed edges). Two nodes in the line graph are connected if the corresponding edges in the primal graph share a node. Notice that, while the number of primal edges (7) is equal to the number of dual nodes (7), the number of dual edges (10) is not necessarily equal to the number of primal nodes (6). (B) Combinatorial optimization as a single player game defined by states, actions, and reward. Traversing a path (green) from the root to a leaf node (pink square) corresponds to a solution for a problem. White nodes represent states and black nodes represent actions. From each state there may be many possible actions (more than the two illustrated here) representing the possible nodes or edges in the problem graph. 
The leaf nodes represent rewards or costs such as the sum of weights in MST or 
length of tour in TSP. 
(C) Graph algorithms for polynomial problems MST and SSP, and NP-hard problems TSP and VRP formulated as single player games by reinforcement learning using states, actions, and reward. 
For MST, the state includes the graph, line graph, weights, and 
selected edges $\mathcal T$ (red).}
\label{fig:unified-framework}
\end{figure}

Given a graph $\mathcal{G}=(V,E,W)$, $V=\{1,\ldots,n\}$ is the set of vertices (nodes), $E$ is the set of edges, and $W$ is the set of edge weights. 
For edges $e_{ij}$ between nodes $i$ and $j$ in an undirected graph $w_{ij}=w_{ji}$. $|V|$ and $|E|$ represent the number of vertices and edges in the graph. Given a node $i$, $\mathcal{N}(i)$ denotes the set of its neighboring nodes.

\paragraph{Line graph.}
Given a primal graph $\mathcal{G}=(V,E,W)$, the edge-to-vertex dual or \emph{line graph}, $\mathcal{G}^*=(V^*,E^*,W^*)$, is defined so each edge in the primal graph 
corresponds to a node in the line graph: $V^*=E$. 
Two nodes in the line graph are connected if the corresponding edges in the primal graph share a node. 
Edge weights in the primal graph become node weights $W^*$ in the line graph.
Figure \ref{fig:unified-framework}(A) illustrates the relationship between the primal and line graphs.

\paragraph{Graph generation.}
We learn 
MST and single-source shortest paths (SSP) by training and running on five different types of random graphs: Erd\H{o}s-R{\'e}nyi (ER) \cite{erdHos1960evolution}, Barab{\'a}si-Albert (BA) \cite{albert2002statistical}, Stochastic block model (SBM) \cite{HOLLAND1983SBM}, Watts-Strogatz (WS) \cite{watts1998collective}, and Random regular (RR) \cite{steger1999generating, kim2003generating}. 
We learn
TSP and the vehicle routing problem (VRP) by training and running on complete graphs with different numbers of random nodes drawn uniformly from $[0, 1]^2$.
For MST and SSP, edge weights are chosen uniformly between 0 and 1 for pairs of nodes that are connected. 
For TSP and VRP, these weights are the distances between the nodes.
We also test our models on real-world graphs \cite{tsplib}.

\subsection{Problems over Graphs}

\paragraph{Minimum Spanning Tree (MST).}
Given a connected and undirected graph $\mathcal{G}=(V,E,W)$, the MST problem is to find a tree $\mathcal{T}=(V_\mathcal{T},E_\mathcal{T})$ with $V_\mathcal{T} = V$, $E_\mathcal{T}\subset E$ minimizing the sum of the edge weights $W_\mathcal{T}\subset W$.
Algorithms for MST problems include Boruvka's \cite{nevsetvril2001otakar}, Prim's \cite{prim1957shortest} and Kruskal's algorithm \cite{kruskal1956shortest}; all are greedy algorithms with time complexity $O(|E|\log|V|)$.

\paragraph{Single-Source Shortest Paths (SSP).}
We consider the SSP problem with non-negative edge weights. Given a connected and directed graph $\mathcal{G}=(V,E,W)$ and a source vertex, the single-source shortest paths problem is to find the shortest paths from the source to all other vertices.
For the single-source shortest paths problem with nonnegative weights, Dijkstra's algorithm \cite{dijkstra1959note} complexity is $O(|V|\log |V| + |E|)$ using a heap. 
For the general single-source shortest paths problem, Bellman-Ford \cite{bang2000section} runs in $O(|V||E|)$. 
In addition, the Floyd–Warshall algorithm \cite{cormen1990introduction} solves the shortest paths problem between all pairs of nodes with cubic time complexity $O(|V|^3)$.

\paragraph{Traveling Salesman Problem (TSP).}
Given a graph $\mathcal{G}=(V,E,W)$, 
let $V$ represent a list of cities and $W$ represent the distances between each pair of cities. 
The goal of TSP is to find the shortest tour that visits each city once and returns to the starting city. 
TSP is an NP-hard problem. 
Approximation algorithms and heuristics include LKH \cite{lin1973effective}, Christofides \cite{christofides1976worst}, 2-opt \cite{lin1965computer,aarts2003local}, Farthest Insertion and Nearest Neighbor \cite{rosenkrantz1977analysis}.
Concorde \cite{applegate2006concorde} is an exact TSP solver.
Gurobi \cite{gurobi2019} is a general integer programming solver which can also used to find an exact TSP solution.

\paragraph{Vehicle Routing Problem (VRP).}
Given $M$ vehicles and a graph $\mathcal{G}=(V,E)$ with $|V|$ cities, 
the goal of the VRP is to find optimal routes for the vehicles. 
Each vehicle $m\in\{1,..,M\}$ starts from the same depot node, 
visits a subset $V(m)$ of cities, and returns to the depot node. 
The routes of different vehicles do not intersect except at the depot; 
together, the vehicles visit all cities. 
The optimal routes minimize the longest tour length of any single route. 
TSP is a special case of VRP for one vehicle.

\subsection{Learning Graph Algorithms as Single Player Games}

We represent the problem space as a search tree. The leaves of the search tree represent all (possibly exponentially many) possible solutions to the problem. A search traverses this tree, choosing a path guided by a neural network as shown in Figure \ref{fig:unified-framework}(B). The initial state, represented by the root node, may be the empty set, a random state, or other initial state. Each path from the root to a leaf consists of moving between nodes (states) along edges (taking actions) reaching a leaf node (reward). Actions may include adding or removing a node or edge. The reward (or cost) may be the value of the solution, for example a sum of weights or length of tour. For each problem, Figure \ref{fig:unified-framework}(C) defines the states, actions, and reward within our framework. In the supplementary material, we show that the single player formulation extends our framework to other combinatorial optimization problems on graphs.

When the predictions of our neural network capture the global structure of the problem, this mechanism is very efficient. On the other hand, even if the network makes poor predictions for a particular problem, the search will still find the solution if run for a sufficiently long (possibly exponential) time. The network is retrained using the results of the evaluation oracle on the leaf nodes reached by the search to improve its predictions. In the context of perfect information games, 
\cite{sun2018dual} proved that a similar mechanism converges asymptotically to the optimal policy.

\section{Methods}

\begin{figure}[b]
\centering
\includegraphics[width=0.7\textwidth]{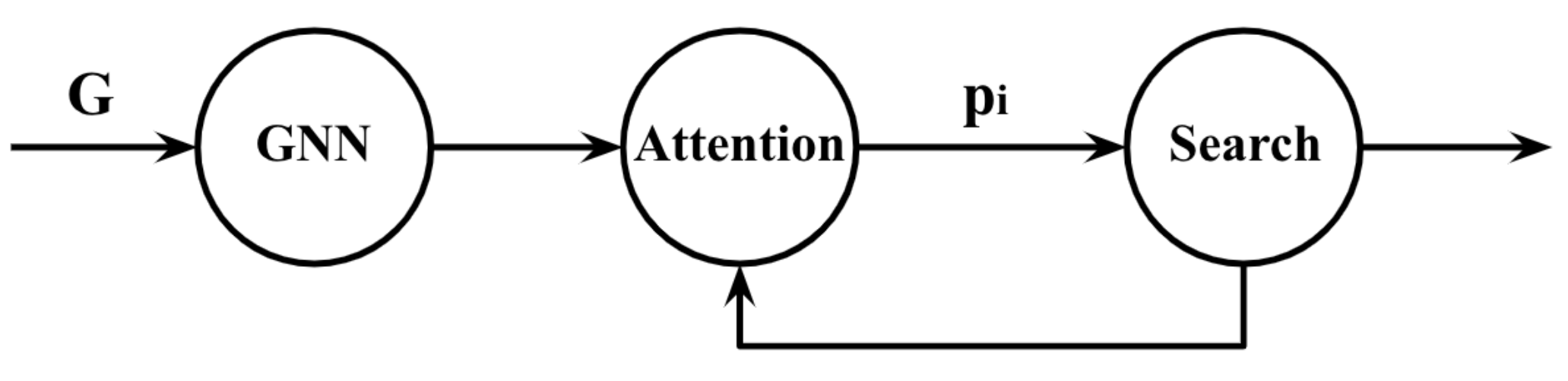}
\caption{Architecture: given a graph for a combinatorial optimization problem, the features (city coordinates for TSP and VRP and edge weights for MST and SSP) of the graph are encoded by a graph neural network. 
The encoded features are passed into an attention mechanism to predict the probabilities over the next nodes/edges. The next node/edge is selected according to this probability.}
\label{fig:model-free-rl-with-gnn}
\end{figure}

We introduce a framework using model-free RL with a GNN representation. The framework is general since solving different problems only requires changing the objective function, whereas the architecture remains the same.  
Our framework uses a GNN to choose nodes. 
We transform problems that require choosing edges, such as MST and SSP, 
into a node problem using the line graph.
Our framework is based on an encoder-decoder architecture, using a graph attention network (GAT) as the encoder and attention model as the decoder. The architecture is illustrated in Figure \ref{fig:model-free-rl-with-gnn}. We begin by computing the line graph, so that the edge weights become node weights (features). 
The primal and line graph features are inputs to the GNN.
For TSP, node coordinates are direct inputs to the GNN.  

\paragraph{Encoder.}
The input is a graph $\mathcal{G}=(V,E,W)$ with node features $\mathbf{v}_i\in\mathbb{R}^d$ for each node $i \in \{1,\ldots,n\}$ where $n = |V|$ is the number of graph nodes. 

The encoder of our framework is a graph attention network (GAT) \cite{velivckovic2017graph}, a type of GNN. Each layer $l=1,...,L$
of the GNN updates the feature vector at the $i$-th node as:
\begin{equation}
\mathbf{v}_i^{l} =\alpha_{ii}\Theta\mathbf{v}_i^{l-1} +\sum_{j\in\mathcal{N}(i)}\alpha_{ij}\Theta\mathbf{v}_j^{l-1},
\end{equation}
where $\Theta^{l}$ is a learnable weight matrix, $\mathcal{N}(i)$ are the neighbors of the $i$-th node, 
and $\alpha^{l}_{ij}$ are the attention coefficients, defined as:
\begin{equation}
\alpha^{l}_{ij} = \frac{\exp(\sigma(z^{l^{\top}}\left[\Theta^{l}\mathbf{v}^{l}_i,\Theta^{l}\mathbf{v}^{l}_j\right]))}{\sum_{k\in\mathcal{N}(i)}\exp(\sigma(z^{l^{\top}}\left[\Theta^{l}\mathbf{v}^{l}_i,\Theta^{l}\mathbf{v}^{l}_k\right]))},
\end{equation}
where $z^{l}$ is a learnable vector, and $\sigma(\cdot)$ is the leaky ReLU activation function. Our GAT consists of three layers $l = 1,2,3$. The last layer $L$ of our GAT is a softmax which computes a vector of probabilities over nodes. 

The network recommends the node that maximizes this probability.

\paragraph{Decoder.}
The decoder uses an attention mechanism \cite{kool2019attention}. 
We describe the decoder in the context of the TSP for simplicity.
Suppose at the previous iteration, we selected node $i$ with features $\mathbf{v}_i$ and neighborhood $j \in \mathcal N(i)$.
Next, we compute the attention coefficients
 
$\alpha^\textup{dec}_{ij}$, defined as:
\begin{equation}
\alpha^\textup{dec}_{ij} = C\tanh\left((\Phi_1\mathbf{v}_i)^{\top} (\Phi_2\mathbf{v}_j) /\sqrt{d_h}\right),
\end{equation}

where $\Phi_1$,$\Phi_2\in\mathbb{R}^{d_h\times d}$ are learned parameters,
$d_h$ is the hidden dimension, and $C \in \mathbb{R}$ is a constant.

The probabilities of selecting the next node of the graph are proportional to the attention coefficients using the softmax.

\paragraph{Search.} Efficient search algorithms for combinatorial optimization problems include beam search, neighborhood search, and tree search. 
Here, we use the predicted edge/node probabilities to select the next edge/node in the graph in a greedy fashion. 

\paragraph{Algorithm.}
Given an input graph, the algorithm first computes its line graph (for problems that require selecting edges). 
The node or edge weights 
are used as features for the GNN. 
The algorithm first uses the GAT encoder to create a dense representation of the input graph. 
The algorithm then loops for $n$ iterations, 
using the decoder to select a primal node/edge to add, as an action. 
Loss is calculated using the policy gradient of the rewards. 
Next, we present the solution to each problem within our unified framework. Detailed pseudo-code for our algorithm, along with solutions for SSP and VRP, are included in the supplementary material.

\subsection{Solutions}

\paragraph{MST.} We consider graphs $\mathcal{G}=(V,E,W)$, where $W$ is the set of edge weights. Suppose we select an edge by policy $\pi$.  Then the problem is formulated as finding a subset of edges $E_{\pi}\subset E$ and weights $W_{\pi}\subset W$, such that $\mathcal{T}=(V,E_{\pi},W_{\pi})$ is a tree, and among all trees, the sum of the edge weights $W_\pi$ is minimal. 
The reward is defined as the negative sum of the edge weights:
\begin{equation}
\label{eq:mst-reward}
r = -\left(I(\mathcal{T}) + \sum_{e\in E_{\pi}}W(e)\right),
\end{equation}
where $I$ is an indicator function such that $I(\mathcal{T}) = 0$ if $\mathcal{T}$ is a tree and is a large value otherwise.
To search for an approximate solution for the MST problem we need to select edges. We therefore first compute the dual graph, turning each edge in the primal into a node in the dual.

We define the dual (line) graph $\mathcal{G}^*=(V^*,E^*)$, where $|V^*|=|E|$. The dual node features $\mathbf{v}_{i}$ are the primal edge weights $w_{i}$ for $i \in \{1, \ldots, |E|\}$, such that $\mathbf{v}_{i} = w_{i}$.

We train the GNN to learn node features $\mathbf{v}_{i}^L\in\mathbb{R}^{d_L}$ in the line graph and obtain the probability distribution $\mathbf{p}\in\mathbb{R}^{|E|}$ over edges in the primal graph. Our algorithm iteratively selects edges in the primal and maintains the set of their adjacent nodes $V_{\pi}$, until $V_{\pi} = V$.
At run time, if a state is not a tree then it is penalized by a negative reward. 
Hence the GNN learns to identify trees in addition to identifying low-weight edges.

\paragraph{TSP.}
For the Euclidean TSP problem, we define our node features as a set of 2D coordinates $\mathbf{v}_i\in\mathbb{R}^2$. 
The input to the GNN is the graph $\mathcal G = (V, E)$ and the node features $\mathbf{v}_{1}, \ldots, \mathbf{v}_{N}$.
Our goal is to find a permutation $\tau$ of nodes with the shortest tour length.
In our RL framework, at each iteration $i$,
the state is the set of nodes $\tau(1),\ldots,\tau(i-1)$ visited so far,
and the action is the choice of the next node $\tau(i)$ to visit.
The reward is defined as the negative expected total tour length: 
\begin{equation}
\label{eq:tsp-reward}
r=-\sum_i^{|V|}\|\mathbf{v}_{\tau(i)}-\mathbf{v}_{\tau(i+1)}\|_2,
\end{equation}
where $\tau(|V|+1)=\tau(1)$ represents the starting node.
In each step,  we use the probability distribution $\mathbf{p}_i$ over all unselected nodes to predict the next node to visit. Nodes that have already been selected are masked by setting their probabilities to $0$. TSP is a special case of VRP with a single vehicle. 

\section{Results}

\begin{figure}[ht]
\centering
\subfigure[MST]{
\includegraphics[width=0.49\textwidth]{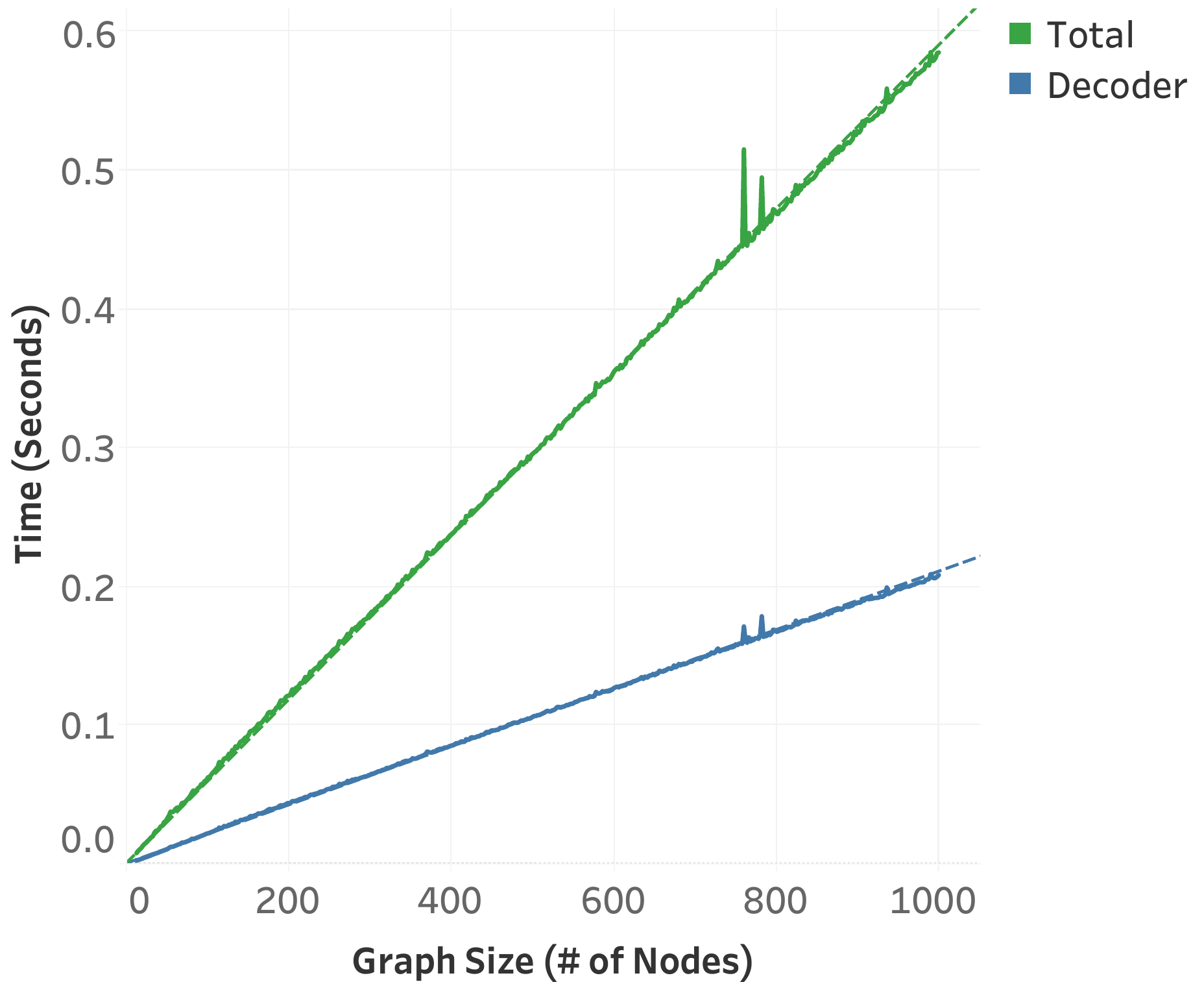}}
\subfigure[TSP]{
\includegraphics[width=0.49\textwidth]{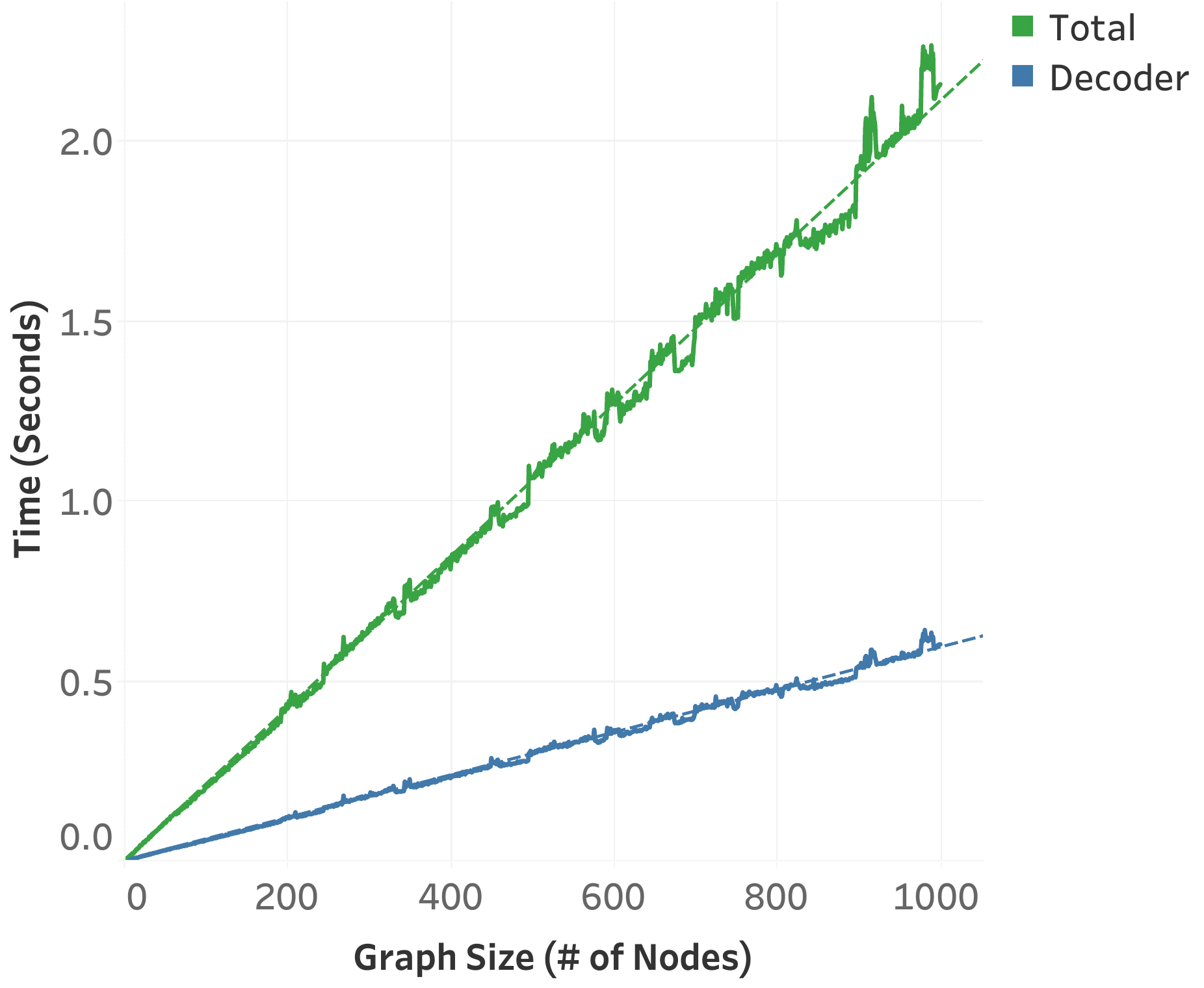}}
\caption{(a) MST and (b) TSP running times of the decoder, and total time for $1000$ graphs of sizes $1$--$1000$.
Each of a $1000$ runs is performed on a single batch. 
The encoder is quite fast and is only run once per graph and has output that is linear in the number of nodes; on these problems the median running time of the encoder is under two milliseconds. 
Our method provides an approximation in linear time $O(n)$ in the number of graph nodes $n$.}
\label{fig:mst-tsp-linear-time}
\end{figure}

\begin{figure}[ht]
\centering
\subfigure[MST Weights]{
\includegraphics[height=6.8cm]{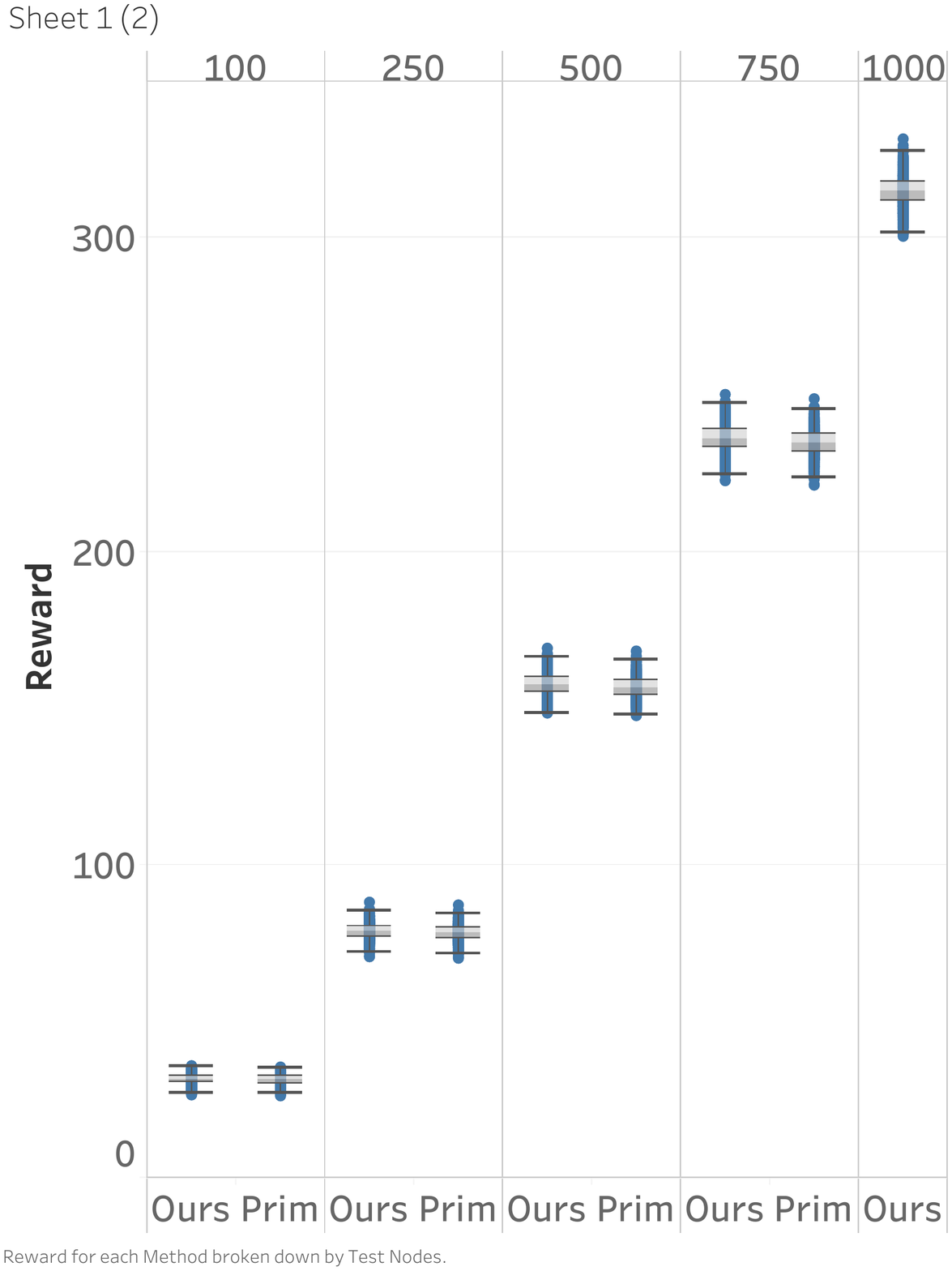}}
\subfigure[MST Opt. Gaps]{
\includegraphics[height=6.8cm]{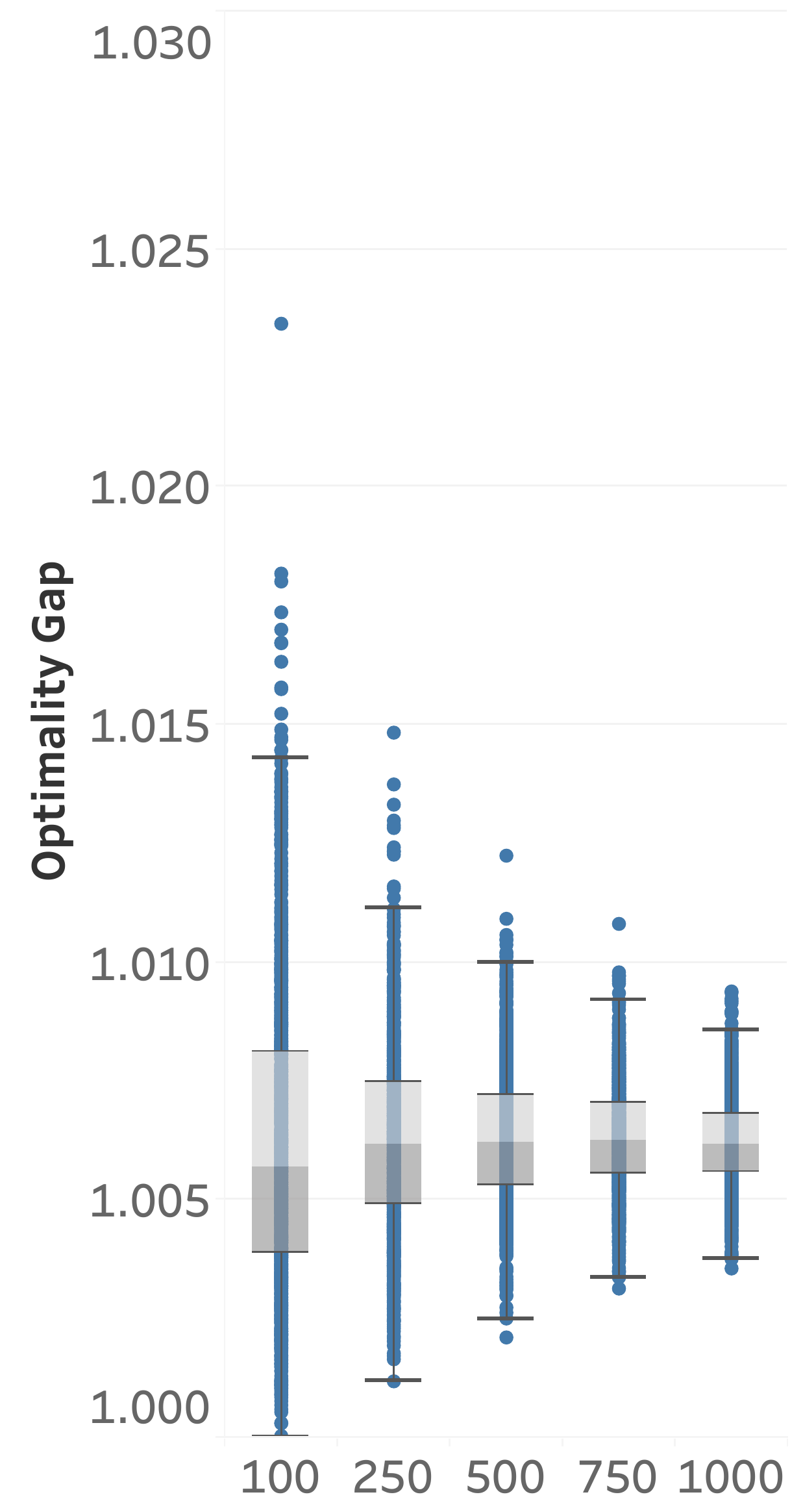}}
\caption{Generalization on MST from small to large random regular graphs: (a) Weights for our method compared with Prim on graphs of increasing size. Our model is trained on graphs with 100 nodes and run on graphs with 100, 250, 500, 750 and 1000 nodes. For each graph size, we run our algorithm and Prim's algorithm on the exact same 1000 graphs. The quantiles increase for both methods as we test on larger graphs. (b) Optimality gaps. As graph size increases, the quantiles in the box-whiskers plot decrease.}
\label{fig:mst-optimality-gap}
\end{figure}

\begin{figure}[ht]
\centering
\includegraphics[width=0.7\textwidth]{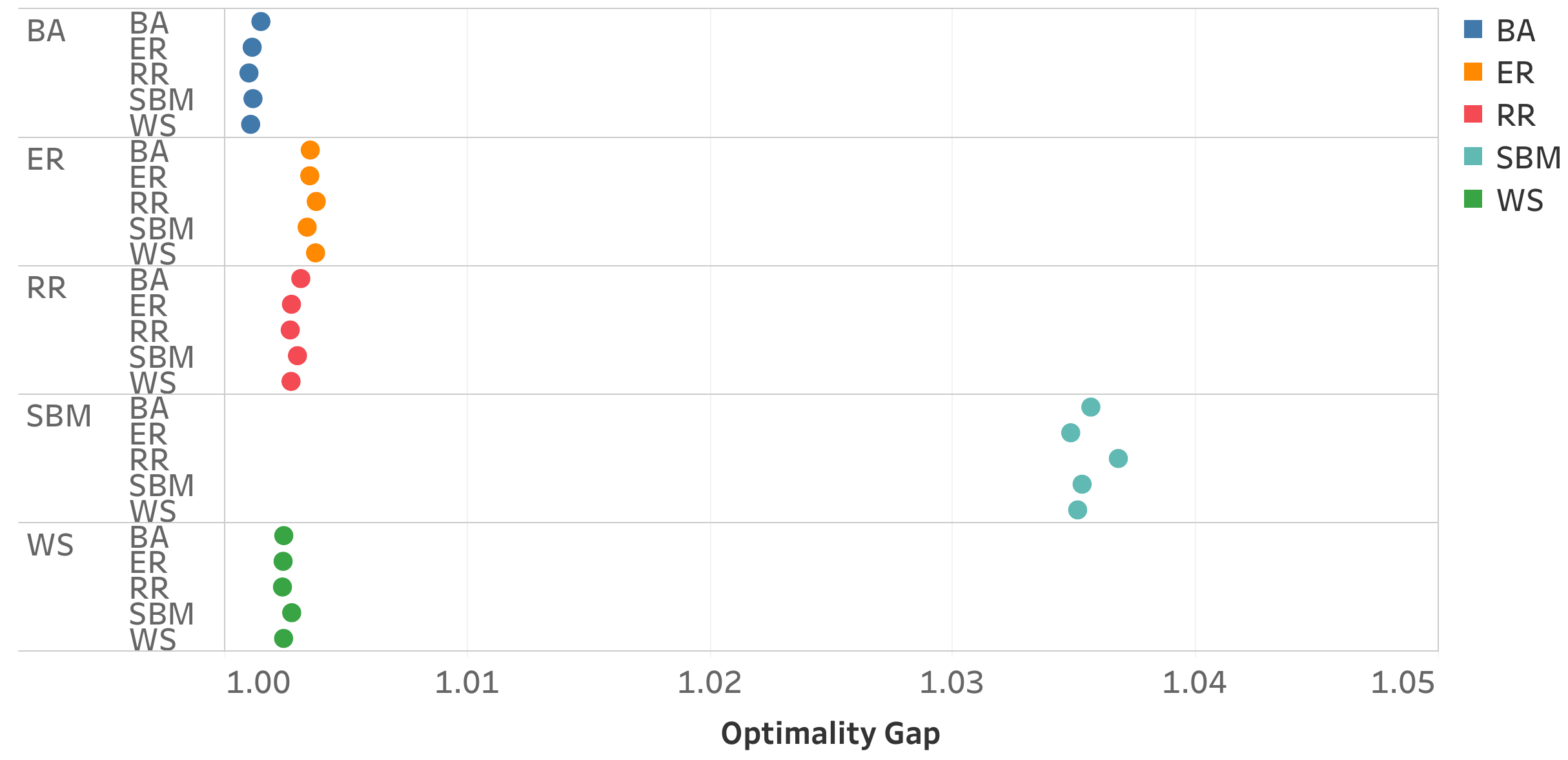}
\caption{Generalization on MST between different random graph types. Values denote optimality gaps over 1000 random graphs with 50 nodes each. Training from different graph types are grouped together by color and leftmost labels. Our algorithm has excellent generalization from BA, ER, RR, and WS random graphs with optimality gaps between $1.001$-$1.0037$, and very good generalization from SBM random graphs with optimality gaps between $1.034$-$1.037$. }
\label{fig:graph-types}
\end{figure}

\begin{figure}[ht]
\centering
\subfigure[TSP Tour Length]{
\includegraphics[height=5.8cm]{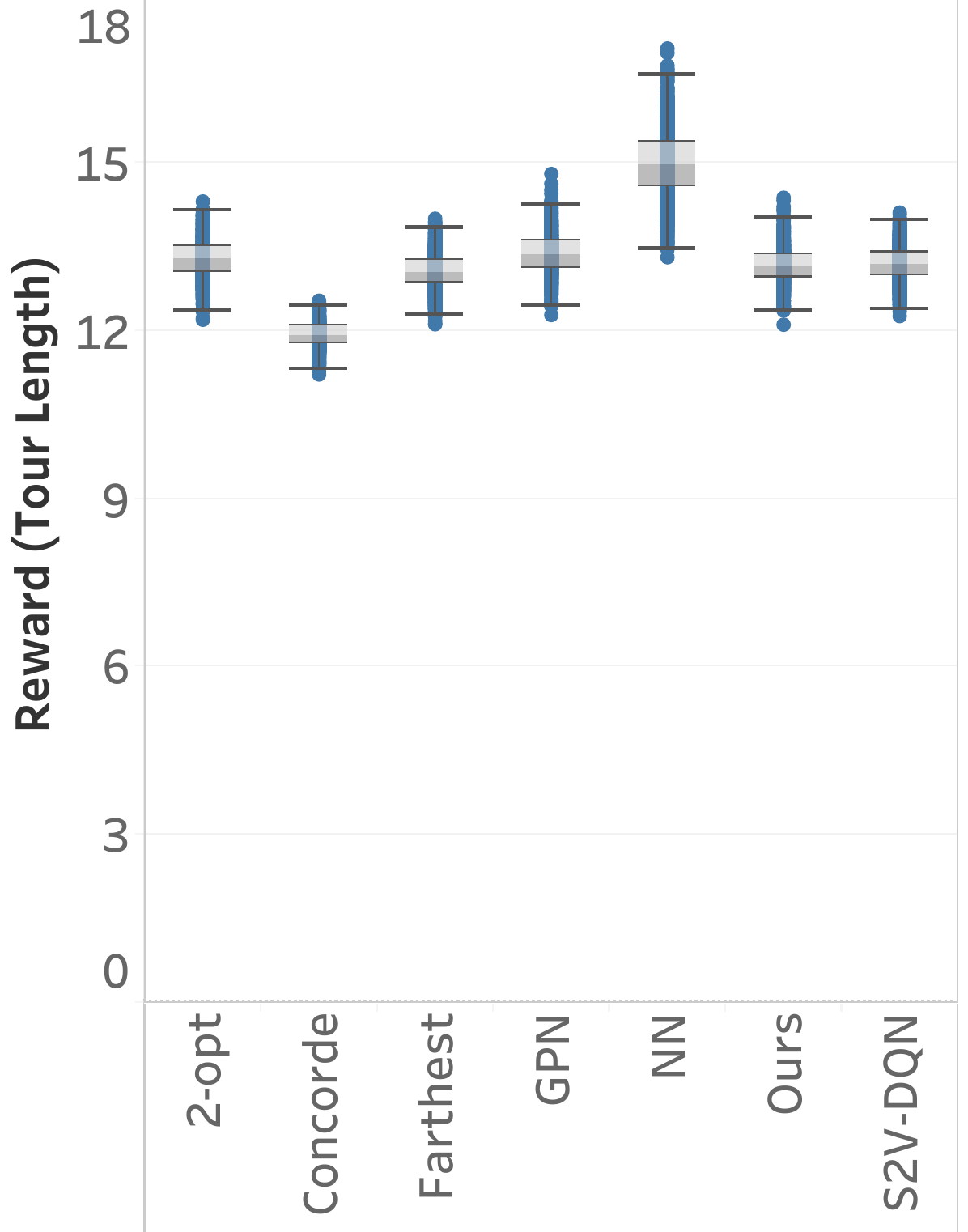}}
\subfigure[TSP Opt. Gaps]{
\includegraphics[height=5.6cm]{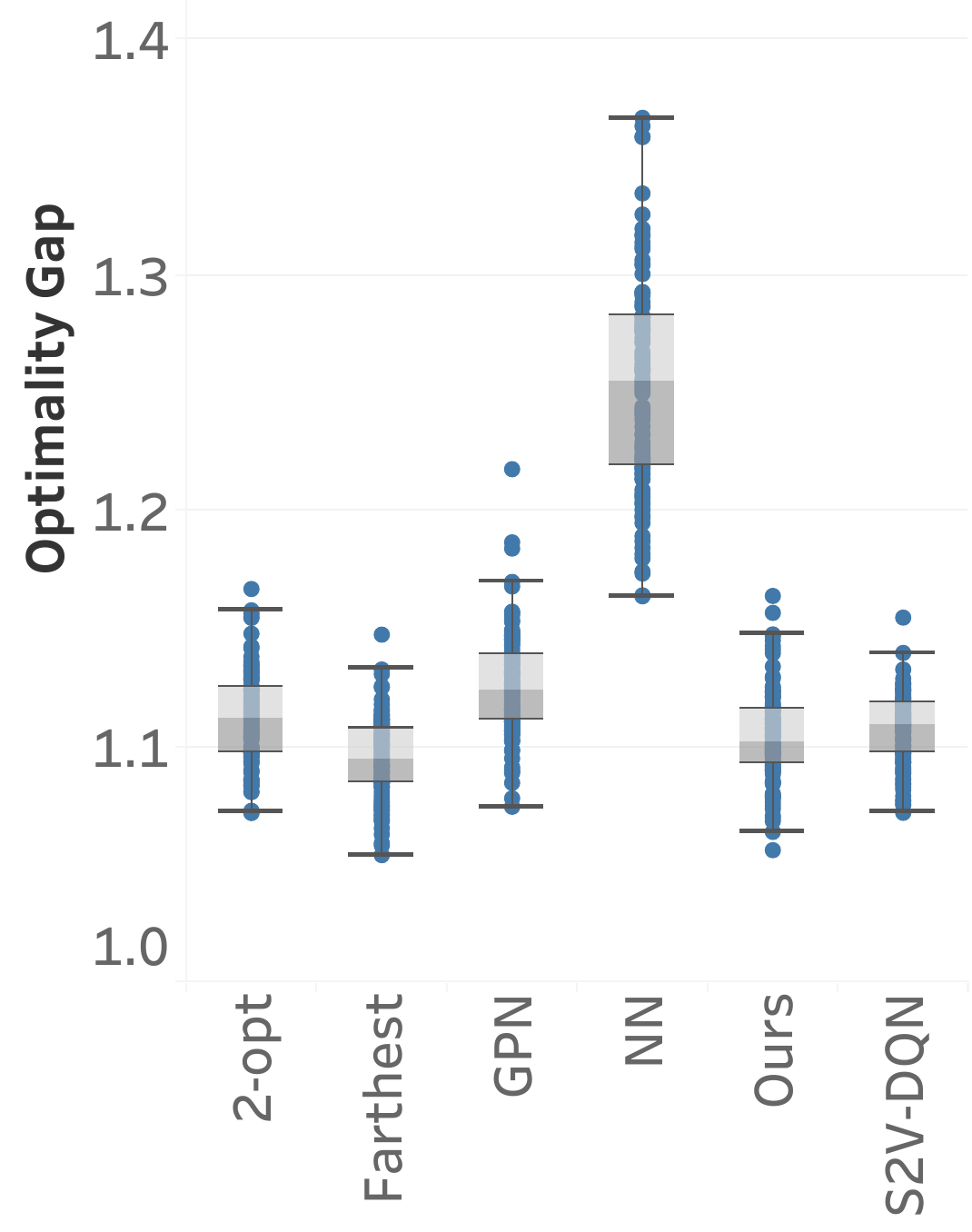}}
\caption{Generalization on TSP from small to large random Euclidean graphs: (a) Comparison of tour lengths of different methods. Our model is trained on graphs with 100 nodes and run on 1000 graphs with 250 nodes. All methods are compared on the same 1000 graphs. (b) Optimality gaps between each method and Concorde which computes the optimal.}
\label{fig:tsp-small2large-optimality-gap}
\end{figure}

Overall, we performed experiments with multiple runs (predicting for ML methods, approximating and running heuristics, or solving for optimality for non-ML methods) on various combinations of problems, methods, number of graph nodes used during training, and number of graph nodes used for running. We performed all experiments on the Google cloud with an NVIDIA Tesla P100 GPU, training for several hours, for each experiment. We compared the performance (reward) and running time on the same graph instances across the various methods.
Figure \ref{fig:mst-tsp-linear-time} shows that our approach has linear running time in the number of graph nodes. 
The left panel shows running time for MST on RR graphs 
and the right panel shows running time for TSP on random Euclidean graphs 
as we increase the number of nodes in the graph from 1 to 1000. 
Figure \ref{fig:mst-optimality-gap} shows that our method converges to optimality for polynomial problems and that our approach generalizes to larger graphs while maintaining an optimality gap close to 1. Figure \ref{fig:graph-types} shows that our approach generalizes from one type of random graph to another while maintaining optimality gaps close to 1. Figure \ref{fig:tsp-small2large-optimality-gap} shows that our method generalizes well from training on small random Euclidean graphs to running on large graphs for TSP. 
Table \ref{tab:tsp-real-world} shows that our method generalizes from random graphs to real-world graphs from TSPLIB \cite{tsplib} while maintaining an optimality gap close to 1. 

\begin{table}[ht]
\centering
\small
\begin{tabular}{l|c|ccc|ccc}
\textbf{TSPLIB} & \textbf{Exact} & & \textbf{RL} & & & \textbf{Approx.} &\\
\textbf{Instance} & \textbf{Concorde} & \textbf{Ours} & \textbf{GPN} & \textbf{S2V-DQN} & \textbf{Farthest} & \textbf{2-opt} & \textbf{Nearest}\\
\hline
eil51 & 426 & \textbf{439} & 485 & \textbf{439} & 448 & 452 & 514\\
berlin52 & 7,542 & \textbf{7,681} & 8,795 & 7,734 & 8,121 & 7,778 & 8,981\\
st70 & 675 & \textbf{684} & 701 & 685 & 729 & 701 & 806\\
eil76 & 538 & \textbf{555} & 591 & 558 & 583 & 597 & 712 \\
pr76 & 108,159 & 112,699 & 118,032 & \textbf{111,141} & 119,649 & 125,276 & 153,462 \\
rat99 & 1,211 & 1,268 & 1,472 & \textbf{1,250} & 1,319 & 1,351 & 1,565 \\
kroA100 & 21,282 & \textbf{21,452} & 24,806 & 22,335 & 23,374 & 23,306 & 26,856 \\
kroB100 & 22,141 & \textbf{22,488} & 24,369 & 22,548 & 24,035 & 23,129 & 29,155 \\
kroC100 & 20,749 & \textbf{21,427} & 24,780 & 21,468 & 21,818 & 22,313 & 26,327 \\
kroD100 & 21,294 & \textbf{21,555} & 23,494 & 21,886 & 22,361 & 22,754 & 26,950 \\
kroE100 & 22,068 & \textbf{22,267} & 23,467 & 22,820 & 23,604 & 25,325 & 27,587 \\
rd100 & 7,910 & \textbf{8,243} & 8,844 & 8,305 & 8,652 & 8,832 & 9,941 \\
eil101 & 629 &  \textbf{650} & 704 & 667 & 687 & 694 & 825 \\
lin105 & 14,379 &  \textbf{14,571} & 15,795 & 14,895 & 15,196 & 16,184 & 20,363 \\
pr107 & 44,303 & 44,854 & 55,087 & \textbf{44,780} & 45,573 & 46,505 &  48,522 \\
pr124 & 59,030 &  \textbf{59,729} & 67,901 & 61,101 & 61,645 & 61,595 & 69,299 \\
bier127 & 118,282 &  \textbf{120,672} & 134,089 & 123,371 & 127,795 & 136,058 & 129,346 \\
ch130 & 6,110 & \textbf{6,208} & 6,457 & 6,361 & 6,655 & 6,667 & 7,575 \\
pr136 & 96,772 & \textbf{98,957} & 110,790 & 100,185 & 104,687 & 103,731 & 120,778 \\
pr144 & 58,537 & 60,492 & 67,211 & \textbf{59,836} & 62,059 & 62,385 & 61,651 \\
ch150 & 6,528 & \textbf{6,729} & 7,074 & 6,913 & 6,866 & 7,439 & 8,195 \\
kroA150 & 26,524 & \textbf{27,419} & 30,260 & 28,076 & 28,789 & 28,313 & 33,610 \\
kroB150 & 26,130 & 27,165 & 29,141 & \textbf{26,963} & 28,156 & 28,603 & 32,825 \\
pr152 & 73,682 & 79,326 & 85,331 & 75,125 & \textbf{75,209} & 77,387 & 85,703 \\
u159 & 42,080 & 43,687 & 52,642 & 45,620 & 46,842 & \textbf{42,976} & 53,637 \\
rat195 & 2,323 & \textbf{2,384} & 2,686 & 2,567 & 2,620 & 2,569 & 2,762 \\
d198 & 15,780 & 17,754 & 19,249 & 16,855 & \textbf{16,161} & 16,705 & 18,830 \\
kroA200 & 29,368 & \textbf{30,553} & 34,315 & 30,732 & 31,450 & 32,378 & 35,798 \\
kroB200 & 29,437 & \textbf{30,381} & 33,854 & 31,910 & 31,656 & 32,853 & 36,982 \\
ts225 & 126,643 & \textbf{130,493} & 147,092 & 140,088 & 140,625 & 143,197 & 152,494 \\
tsp225 & 3,916 & 4,091 & 4,988 & 4,219 & 4,233 & \textbf{4,046} & 4,748 \\
\hline
\textbf{Mean Opt. Gap} & 1 & \textbf{1.032} & 1.144 & 1.045 & 1.074 & 1.087 & 1.238\\
\end{tabular}
\caption{TSP generalization from random graphs to real world graphs, 
comparing tour lengths of different methods. 
TSPLIB \cite{tsplib} instance suffixes denote number of nodes. Concorde computes the optimal tour. 
Mean optimality gap of each method appears in the last row, and our method performs best. 
RL methods GPN, S2V-DQN, and ours are trained on random Euclidean graphs with 100 nodes.}
\label{tab:tsp-real-world}
\end{table}

\section{Conclusions}
In this work we provide a unified framework using reinforcement learning with a GNN representation for learning to approximate different problems over graphs. Our framework is the first to use the line graph for operating on either nodes or edges, and solves different problems by changing the reward function. We present comprehensive studies of generalization from small to large random graphs, between different types of random graphs, and from random to real-world graphs. Finally, we demonstrate linear running times in both theory and practice, running orders of magnitudes faster than other methods with very good optimality gaps. In future work we plan on learning to decompose extremely large instances into parts, solve each part, and combine the partial solutions into a whole. We also plan on meta-learning combinatorial optimization problems over graphs by transfer learning between different reducible problems, learning to learn from the solution of one problem to another.

\subsection{Broader Impact}
Our work has broader implications in the fields of Computer Science and Operations Research for automatically learning algorithms on graphs. Traditionally, approximation algorithms and heuristics for each problem have been developed by computer scientists. In contrast, our unified framework learns graph algorithms without expert knowledge. As a result, our approach eliminates the need to develop approximations and heuristics tailored to each problem. 

We build neural network architectures with linear runtimes by construction. Perhaps most importantly, our results yield optimality gaps close to 1 for real world problems on graphs. In a similar fashion to computer programs that play chess in a league of their own, we envision that our work is a step towards learning graph algorithms that improve and out-perform human algorithms developed over decades in the field.

With an automated approach, it is important to consider the limitations in explainability. Automatically learning combinatorial optimization algorithms on graphs presents a challenge for humans to understand the resulting algorithms which may present a challenge in mission-critical applications. On the other hand, it is generally easy to check the quality of a proposed solution.

\newpage
\clearpage

\setcounter{section}{0}
\renewcommand{\thesection}{\Alph{section}}

\section*{Supplementary Material}

\section{Summary of Related Work}
\begin{figure}[h]
\centering
\includegraphics[width=0.95\textwidth]{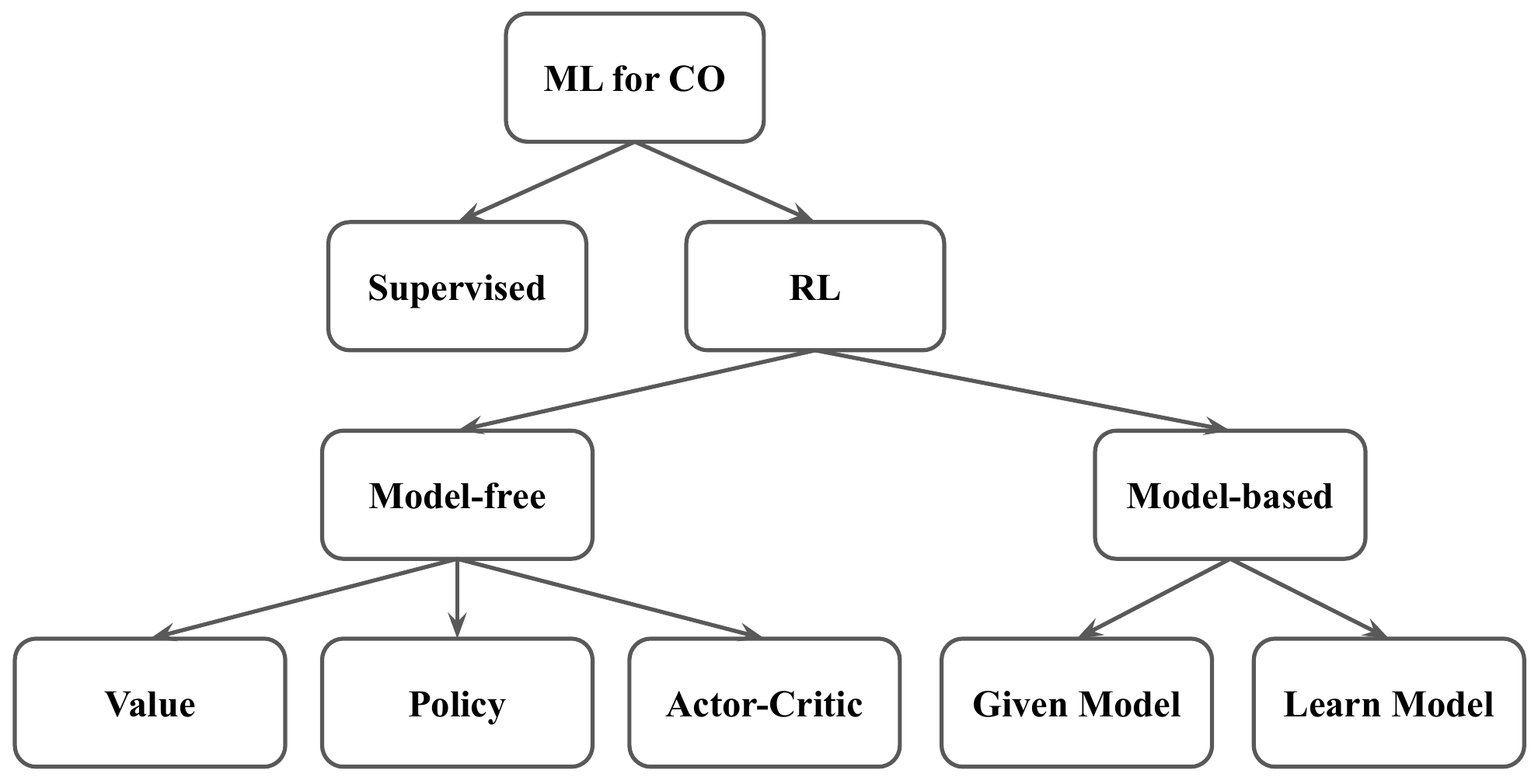}
\caption{Types of ML approaches for NP-hard combinatorial optimization (CO) problems.}
\label{fig:ml4co}
\end{figure}

We include a comprehensive list of supervised and reinforcement learning methods used for combinatorial optimization of NP-hard problems and a classification of all previous work by problem, method, and type, as shown in Table \ref{tab:combopt-work}. Machine learning approaches for these problems may be divided into supervised and reinforcement learning methods as shown in Figure \ref{fig:ml4co}. Reinforcement learning methods can further be divided into model-free and model-based approaches. In turn, model-free approaches may be further divided into value-based, policy-based, or actor-critic methods which are a combination of both. Model-based approaches may further be divided into methods that are given the model and methods that learn the model. Most approaches use a combination of search or reinforcement learning with a graph neural network representation.

As described in Table \ref{tab:combopt-work}, RL using a GNN representation is the most common approach for tackling NP-hard combinatorial optimization problems by ML. The RL setting consists of defining states, actions, and rewards and has been extended to many other problems including Boolean Satisfiability (SAT) \cite{li2018combinatorial}, Maximum Independent Set (MIS) \cite{li2018combinatorial, abe2019solving}, Minimum Vertex Cover (MVC) \cite{hanjun2017learning, li2018combinatorial, song2019co}, Maximum Cut \cite{hanjun2017learning, barrett2019exploratory, beloborodov2020reinforcement}. These problems can be rephrased, or formally reduced, from each other, and may be formulated as a single player game: with states defined as the current solution, actions defined by adding or removing graph nodes or edges, and a reward.

\begin{table*}[t!]
\centering
\small
\begin{tabular}{l|l|l}
    \textbf{NP-hard Problem} & \textbf{Method} & \textbf{Type}\\
    \hline
    Towers of Hanoi & AlphaZero: Recursive MCTS + LSTM \cite{pierrot2019learning} & Model-based, Given model\\
    \hline
    Integer Programming & RL + LSTM \cite{tang2019reinforcement} & Model-free, Policy-based\\
    \hline
    Minimum Dominating Set (MDS)
    & RL + Decision Diagram \cite{cappart2019improving} & Model-free, Value-based\\
    & RL + GNN \cite{yolcu2019learning} & Model-free, Policy-based\\
    \hline
    Maximum Common Subgraph (MCS) & DQN + GNN \cite{bai2020fast} & Model-free, Value-based\\
    \hline
    Maximum Weight Matching (MWM) & DDPG \cite{emami2018learning} & Model-free, Policy-based\\ 
    \hline
    Boolean Satisfiability (SAT) 
    & MPNN \cite{selsam2018learning} & Supervised, Approximation\\
    & RL + GNN \cite{yolcu2019learning} & Model-free, Policy-based\\
    & Tree search + GCN \cite{li2018combinatorial} & Model-based, Given model\\
    \hline
    Graph Coloring
    & RL + GNN \cite{yolcu2019learning} & Model-free, Policy-based\\
    & AlphaZero: MCTS + LSTM \cite{huang2019coloring} & Model-based, Given model\\
    \hline
    Maximum Clique (MC) 
    & RL + GNN \cite{yolcu2019learning} & Model-free, Policy-based\\
    & Tree search + GCN \cite{li2018combinatorial} & Model-based, Given model\\
    \hline
    Maximum Independent Set (MIS) & Tree search + GCN \cite{li2018combinatorial} & Model-based, Given model\\
                                  & AlphaZero: MCTS + GCN \cite{abe2019solving} & Model-based, Given model\\
    \hline
    Minimum Vertex Cover (MVC) 
    & Q-Learning + GNN \cite{hanjun2017learning} & Model-free, Value-based\\
    & DQN, Imitation learning \cite{song2019co} & Model-free, Value-based\\
    & RL + GNN \cite{yolcu2019learning} & Model-free, Policy-based\\
    & Tree search + GCN \cite{li2018combinatorial} & Model-based, Given model\\
    \hline
    Maximum Cut (MaxCut) & Q-Learning + GNN \cite{hanjun2017learning} & Model-free, Value-based\\
                         & DQN + MPNN \cite{barrett2019exploratory} & Model-free, Value-based\\
                         & PPO + CNN, GRU \cite{beloborodov2020reinforcement} & Model-free, Actor-Critic\\
    \hline
    Traveling Salesman Problem (TSP) 
    & Pointer network \cite{vinyals2015pointer} & Supervised, Approximation\\
    & GCN + Search \cite{joshi2019efficient} & Supervised, Approximation\\
    & Q-Learning + GNN \cite{hanjun2017learning} & Model-free, Value-based\\
    & Hierarchical RL + GAT \cite{ma2020combinatorial} & Model-free, Policy-based\\
    & REINFORCE + LSTM with attention \cite{nazari2018reinforcement} & Model-free, Policy-based\\
    & REINFORCE + attention \cite{deudon2018learning} & Model-free, Policy-based\\
    & RL + GAT \cite{kool2019attention} & Model-free, Policy-based\\
    & DDPG \cite{emami2018learning} & Model-free, Policy-based\\ 
    & REINFORCE + Pointer network \cite{bello2016neural} & Model-free, Policy-based\\
    & RL + NN \cite{malazgirt2019tauriel} & Model-free, Actor-Critic \\
    & RL + GAT \cite{cappart2020combining} & Model-free, Actor-Critic\\
    & AlphaZero: MCTS + GCN \cite{parker2020alphatsp} & Model-based, Given model\\
    \hline
    Knapsack Problem & REINFORCE + Pointer network \cite{bello2016neural} & Model-free, Policy-based\\
    \hline
    Bin Packing Problem (BPP) & REINFORCE + LSTM \cite{hu2017solving} & Model-free, Policy-based\\
                              & AlphaZero: MCTS + NN \cite{laterre2018ranked} & Model-based, Given model\\
    \hline
    Job Scheduling Problem (JSP) & RL + LSTM \cite{chen2019learning} & Model-free, Actor-Critic\\
    \hline
    Vehicle Routing Problem (VRP) & REINFORCE + LSTM with attention \cite{nazari2018reinforcement} & Model-free, Policy-based\\
                                  & RL + LSTM \cite{chen2019learning} & Model-free, Policy-based\\
                                  & RL + GAT \cite{kool2019attention} & Model-free, Policy-based\\
                                  & RL + NN \cite{lu2020vrp} & Model-free, Policy-based\\
                                  & RL + GAT \cite{gao2020learn} & Model-free, Actor-Critic\\
    \hline
    Global Routing & DQN + MLP \cite{liao2020deep} & Model-free, Value-based\\
    \hline
    Highest Safe Rung (HSR) & AlphaZero: MCTS + CNN \cite{xu2019learning} & Model-based, Given model\\
    \hline
\end{tabular}
\caption{Classification of ML approaches for NP-hard combinatorial optimization by problem, method, and type.}
\label{tab:combopt-work}
\end{table*}

\subsection{Polynomial Problems and GNN's}
GNNs are equivalent to polynomial algorithms, for specific choices of aggregation and pooling functions. \cite{xu2020reason}. Polynomial problems can be solved using GNNs without RL or search, which may be considered as a fast type-1 process; whereas NP-hard problems require RL or search, which are akin to a slow type-2 process \cite{kahneman2011thinking} as illustrated in Figure \ref{fig:conclusion}. For example, breadth-first search (BFS) and MST have been solved using a message passing neural network (MPNN) \cite{velikovi2020neural} and SSP has been solved using both a GNN \cite{battaglia2018relational, xu2020reason} and MPNN \cite{velikovi2020neural}. However, in all these cases, the methods were given the graph algorithm as expert knowledge. In contrast, our unified framework: (i) is applicable to both polynomial and NP-hard problems; and (ii) does not bake in the underlying algorithms and learns them by example, as described next.

\section{Generating Valid Solutions}
\begin{figure}[t]
\centering
\includegraphics[width=0.3\textwidth]{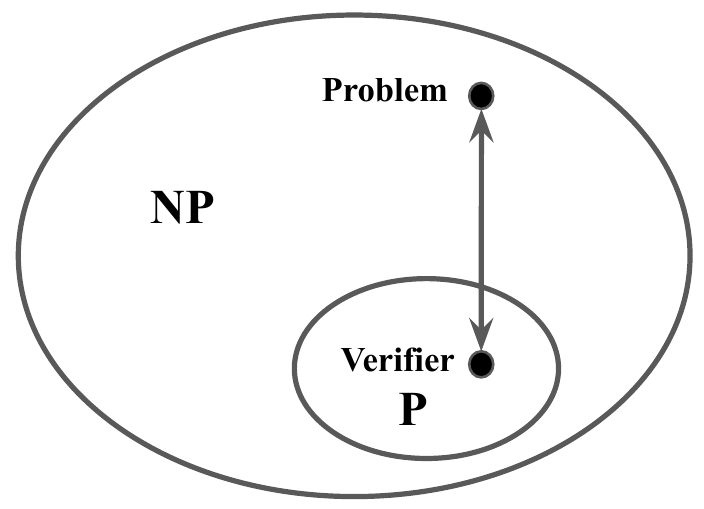}
\caption{Relationship among NP and P. By definition, verification problems for NP-hard problems have polynomial time complexity. Polynomial problems on graphs can be solved using GNNs without RL or search, which may be considered as a fast type-1 process; whereas NP-hard problems require RL or search, which are akin to a slow type-2 process. A GNN can be used directly for verification.}
\label{fig:conclusion}
\end{figure}

Each combinatorial optimization problem has a corresponding verification problem that verifies the solution structure with reduced time complexity compared to the optimization problem. By definition, verification problems for NP-hard problems have polynomial time complexity as illustrated in Figure \ref{fig:conclusion}. We define a validation term in our reward using the verification problem for corresponding optimization problem. We focus on approximate solutions therefore our goals are two-fold: (i) an optimality gap close to 1; and (ii) a valid solution. We therefore model the reward as a combination of the approximation quality and validation of the solution, within a model-free policy-based approach: directly predicting subsequent action probabilities $p(a | s)$ given the state. A solution is validated by using reward penalties based on the verification problem.
The reward function for each problem consists of two terms: an optimization term and a validity term. The optimization term defines the weight or size of the output of the algorithm, such as the weight of an MST tree. The validity term enforces the structure of the solution by penalizing invalid outputs. For example, the validity term of MST penalizes solutions that are not spanning trees. Different problems may have different types of outputs, for example a set of edges for MST and a permutation of nodes for TSP. The line graph handles both nodes and edges in a unified framework. By definition, a permutation is constrained, and therefore may make the validity term redundant for problems whose output is a permutation.

\begin{figure}[h]
\centering
\includegraphics[width=0.5\textwidth]{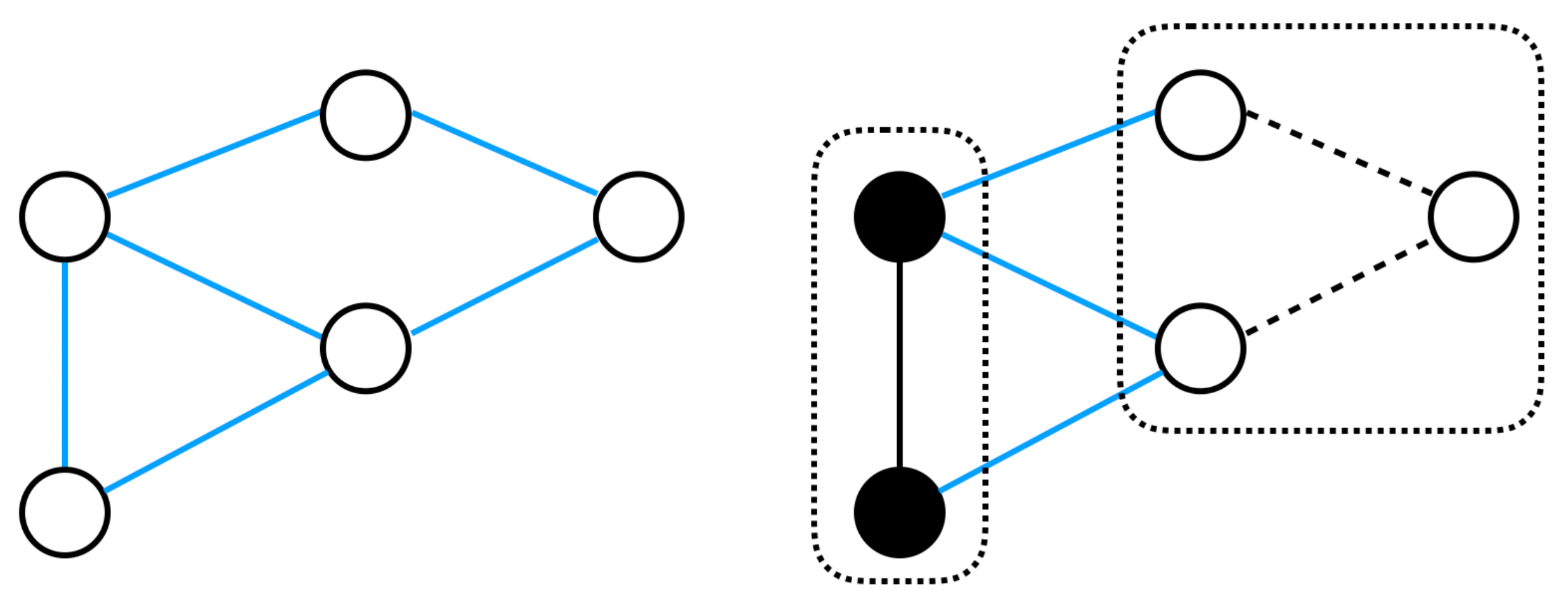}
\caption{Validation for MST: On the left, all edges are available. On the right, we have already selected an edge (and nodes in black) and only (blue) edges connecting black and white nodes can be selected in the next step.}
\label{fig:mst-search}
\end{figure}

One alternative to learning validity within the reward is to model both terms separately using an actor-critic approach: predicting both action probabilities and validity of the next state $v_P(s')$ for a problem $P$, where the encoder-decoder network acts as the policy network. This approach more elegantly evaluates both action and state values, but increases running time. A second alternative is to guarantee validity by an algorithm; however that would incorporate domain knowledge of our problem. For example, for MST, to guarantee that $(V,E_{\pi},W_{\pi})$ is a tree, we would first need to select an initial edge and add its two adjacent nodes into $V_{\pi}$. Then, the next selected edge in the primal must connect two nodes such that one is in $V_{\pi}$ and the other is not, as illustrated in Figure \ref{fig:mst-search}. There exists a randomized
MST algorithm that runs in linear expected time \cite{karger1995randomized} as well as a linear-time algorithm to verify a spanning tree \cite{king1997simpler}. Handling polynomial problems also fits within our unified framework.

\section{Linear Time Approximation Algorithm}
In this section we provide detailed pseudo-code as shown in Algorithm \ref{algorithm} and demonstrate that the worst case running time complexity of our approximation algorithm is linear in the number of nodes or edges. The top section of the algorithm pseudo-code is executed when running the algorithm and has linear time complexity; whereas the bottom section is used only during training.

\paragraph{Line graph.} Given a problem over a graph that requires selecting edges, the algorithm first computes its line graph. Computing the line graph has linear time complexity in the number of edges.

\paragraph{Encoder.} The encoder is run once for each new graph using node or edge weights as features. The encoder is a GAT and has linear running time complexity by construction. The attention layers of the encoder have fixed-size output, and there are a fixed number of attention heads in each GAT layer. In our experiments, the encoder takes a constant time in practice of under 2 milliseconds (which is negligible compared with other steps of the algorithm).

\paragraph{Decoder.} Next, our algorithm loops for $n$ iterations, 
using the decoder to select a primal node/edge to add, as an action. The decoder has a constant running time by considering only non-zero attention coefficients with a constant read-out dimension.

\paragraph{Select.} The selection process considers the action with maximum probability over a constant number of values.

\paragraph{Training loss.} Finally, during training we compute a loss using the policy gradient of the rewards, and update the policy network weights.

\begin{algorithm}[t]
   \caption{}
   \label{algorithm}
\begin{algorithmic}
   \STATE {\bfseries Input:} Graph with features $\mathbf{v}_1,\ldots,\mathbf{v}_{n}$
   \STATE {\bfseries Output:} Approximate solution defined by a set $\mathcal{T}$ or permutation $\tau$
   \STATE Compute line graph (if used then replace $\mathbf{v}_i$'s with $\mathbf{v}_i^*$'s)
   \STATE $x = \text{Encoder} (\mathbf{v}_1,\ldots,\mathbf{v}_{n})$
   \FOR{$t=1$ {\bfseries to} $n$}
    \STATE Compute probabilities $\mathbf{p}=\text{Decoder}\left(x, \mathbf{v}_{\tau(t-1)},\{\mathbf{v}_j\}_{j \in \mathcal{N}(\tau(t-1))}\right)$
    \STATE Take action $a_t = \text{Select} (\mathbf{p})$, updating $\mathcal{T}$ or  $\tau(t)$
    \STATE Compute reward $r_t$
   \ENDFOR
   \\\hrulefill
   \\
   \IF{training}
    \STATE Compute baseline $b$
    \STATE Compute total loss $\mathcal{L}(\theta) =\sum_{t=1}^{n} (r_t-b)\log p(a_t)$
    \STATE Update $\theta= \theta-\nabla \mathcal{L}(\theta)$
   \ENDIF
   \STATE \textbf{return} $\mathcal{T}$ or $\tau$
\end{algorithmic}
\end{algorithm}

\subsection{Solutions for SSP and VRP}
\paragraph{SSP.}
Given a directed graph $\mathcal{G}=(V,E,W)$ and a source node, we assume that there exist paths from the source node to all other nodes. 
We learn a policy $\pi$ such that the sub-graph $\mathcal{G}_{\pi}=(V,E_{\pi},W_{\pi})$ includes the shortest paths $Q_{\pi}$ from the source node to all other nodes. Each path $Q_{i} \in Q_{\pi}$ has a weight $\sum_{(j,k)\in Q_{i}}w_{jk}$. 
The reward is defined as the negative length of all paths:
\begin{equation}
r = -\sum_{i=1}^{|V|}\left(I(\mathcal{Q}_{i})+\sum_{e\in \mathcal{Q}_i}W(e)\right)
\end{equation}
Similar to MST, 
the SSP problem requires us to choose edges, rather than nodes;
therefore we first compute the line graph $\mathcal{G}^*=(V^*,E^*)$ taking node features $\mathbf{v}_i=w_i$ as input to the GNN.

\paragraph{VRP.}
Given $M$ vehicles starting at the depot, we search the graph $M$ times to find $M$ routes. 
For the $m$-th search, the vehicle stops (i.e. we stop searching) if the next selected node is the depot, and then we begin the $(m+1)$-th search. The RL setting is similar to that of TSP except for the definition of reward. The maximum length route of all $M$ routes is minimized, by defining the reward as:
\begin{equation}
\label{eq:vrp-reward}
r=-\max_{m \in \{1, \ldots, M\}} \left\{ \sum_{i \in V_\tau (m)}\|\mathbf{v}_{\tau_m(i)}-\mathbf{v}_{\tau_m(i+1)}\|_2 \right\},
\end{equation}
where $V_{\pi}(m)$ is the set of vertices in the $m$-th route, and both the start and end nodes are the depot.

\section{Generalization Results on Large Real-World Graphs}
Table \ref{tab:tsp-real-world-larger} shows that given sufficient training our approach generalizes well from small random graphs to  large real-world graphs from TSPLIB \cite{tsplib}. Our approach maintains a mean optimality gap below 1.1, which is best among other RL methods and approximation algorithms, while running time is orders of magnitudes faster than other approximation algorithms.

\subsection{Implementation Details}
Our framework consists of an encoder with three multiheaded attention layers and a decoder with an attention mechanism. Layer normalization and ReLU activations are used in the encoder and softmax activations are used in the attention layers. Both the encoder and decoder are updated using the Adam optimizer with a starting learning rate of 1e-3 and 1e-4, respectively, and is reduced by a factor of 10 once learning stagnates.

Overall, we performed millions of runs: predictions using ML methods; approximations, running heuristics, and solving for optimality for non-ML methods, experimenting with various combinations of problems, methods, number of graph nodes for training, and number of graph nodes for running. For example, training on 100 node graphs: MST is trained on 12.8K random graphs with 100 nodes and 250 edges with a batch size of $B=128$ for 100 epochs. TSP is trained on 20M Euclidean random graphs with 100 nodes each for 20K epochs with 16 steps per epoch and batch size $B=64$. Training time for MST is around 90 minutes and training time for TSP is around 48 hours, both on a Google cloud instance with an NVIDIA Tesla P100 GPU.

\begin{table}[t!]
\centering
\small
\begin{tabular}{l|c|ccc|ccc}
\textbf{TSPLIB} & \textbf{Exact} & & \textbf{RL} & & & \textbf{Approx.} &\\
\textbf{Instance} & \textbf{Concorde} & \textbf{Ours} & \textbf{GPN} & \textbf{S2V-DQN} & \textbf{Farthest} & \textbf{2-opt} & \textbf{Nearest}\\
\hline
pr226 & 80,369 & 86,438 & 85,186 & \textbf{82,869} & 84,133 & 85,306 & 94,390\\ 
gil262 & 2,378 & \textbf{2,523} & 5,554 & 2,539 & 2,638 & 2,630 & 3,218\\ 
pr264 & 49,135 & \textbf{52,838} & 67,588 & 53,790 & 54,954 & 58,115 & 58,634\\ 
a280 & 2,579 & \textbf{2,742} & 3,019 & 3,007 & 3,011 & 2,775 & 3,311\\ 
pr299 & 48,191 & 53,371 & 68,011 & 55,413 & 52,110 & \textbf{52,058} & 61,252\\ 
lin318 & 42,029 & \textbf{45,115} & 47,854 & 45,420 & 45,930 & 45,945 & 54,034\\ 
rd400 & 15,281 & 16,730 & 17,564 & 16,850 & 16,864 & \textbf{16,685} & 19,168\\ 
fl417 & 11,861 & 13,300 & 14,684 & \textbf{12,535} & 12,589 & 12,879 & 15,288\\ 
pr439 & 107,217 & 126,849 & 137,341 & 122,468 & 122,899 & \textbf{111,819} & 131,258\\ 
pcb442 & 50,778 & \textbf{55,750} & 58,352 & 59,241 & 57,149 & 57,684 & 60,242\\ 
\hline
\textbf{Mean Opt. Gap} & 1 & 1.095 & 1.331 & 1.106 & 1.105 & 1.096 & 1.252\\
\hline
\end{tabular}
\vspace{10pt}
\caption{TSP generalization from small random graphs to large real world graphs, 
comparing tour lengths of different methods. 
TSPLIB \cite{tsplib} instance suffixes denote the number of nodes in the graph. Concorde computes the optimal tour. 
The mean optimality gap of each method appears in the last row. Our method performs best with running time orders of magnitudes faster than other approximation algorithms. RL methods GPN, S2V-DQN, and ours are trained on random Euclidean graphs with 100 nodes.}
\label{tab:tsp-real-world-larger}
\end{table}

\newpage
\clearpage

\bibliographystyle{plain}
\bibliography{main}

\begin{thebibliography}{10}

\bibitem{aarts2003local}
Emile Aarts, Emile~HL Aarts, and Jan~Karel Lenstra.
\newblock {\em Local Search in Combinatorial Optimization}.
\newblock Princeton University Press, 2003.

\bibitem{abe2019solving}
Kenshin Abe, Zijian Xu, Issei Sato, and Masashi Sugiyama.
\newblock Solving {NP}-hard problems on graphs by reinforcement learning
  without domain knowledge.
\newblock {\em arXiv preprint arXiv:1905.11623}, 2019.

\bibitem{albert2002statistical}
R{\'e}ka Albert and Albert-L{\'a}szl{\'o} Barab{\'a}si.
\newblock Statistical mechanics of complex networks.
\newblock {\em Reviews of {M}odern {P}hysics}, 74(1):47, 2002.

\bibitem{anthony2017thinking}
Thomas Anthony, Zheng Tian, and David Barber.
\newblock Thinking fast and slow with deep learning and tree search.
\newblock In {\em Advances in Neural Information Processing Systems}, pages
  5360--5370, 2017.

\bibitem{applegate2006concorde}
David Applegate, Ribert Bixby, Vasek Chvatal, and William Cook.
\newblock Concorde {TSP} {S}olver, 2006.

\bibitem{bai2020fast}
Yunsheng Bai, Derek Xu, Alex Wang, Ken Gu, Xueqing Wu, Agustin Marinovic,
  Christopher Ro, Yizhou Sun, and Wei Wang.
\newblock Fast detection of maximum common subgraph via deep {Q}-learning.
\newblock {\em arXiv preprint arXiv:2002.03129}, 2020.

\bibitem{bang2000section}
Jrgen Bang-Jensen and Gregory Gutin.
\newblock Section 2.3. 4: The {B}ellman-{F}ord-{M}oore algorithm.
\newblock {\em Digraphs: Theory, Algorithms and Applications}, 2000.

\bibitem{barrett2019exploratory}
Thomas~D Barrett, William~R Clements, Jakob~N Foerster, and AI~Lvovsky.
\newblock Exploratory combinatorial optimization with reinforcement learning.
\newblock {\em In Proceedings of the AAAI Conference on Artificial
  Intelligence}, 2020.

\bibitem{battaglia2018relational}
Peter~W Battaglia, Jessica~B Hamrick, Victor Bapst, Alvaro Sanchez-Gonzalez,
  Vinicius Zambaldi, Mateusz Malinowski, Andrea Tacchetti, David Raposo, Adam
  Santoro, Ryan Faulkner, et~al.
\newblock Relational inductive biases, deep learning, and graph networks.
\newblock {\em arXiv preprint arXiv:1806.01261}, 2018.

\bibitem{bello2016neural}
Irwan Bello, Hieu Pham, Quoc~V Le, Mohammad Norouzi, and Samy Bengio.
\newblock Neural combinatorial optimization with reinforcement learning.
\newblock {\em arXiv preprint arXiv:1611.09940}, 2016.

\bibitem{beloborodov2020reinforcement}
Dmitrii Beloborodov, AE~Ulanov, Jakob~N Foerster, Shimon Whiteson, and
  AI~Lvovsky.
\newblock Reinforcement learning enhanced quantum-inspired algorithm for
  combinatorial optimization.
\newblock {\em arXiv preprint arXiv:2002.04676}, 2020.

\bibitem{bengio2018machine}
Yoshua Bengio, Andrea Lodi, and Antoine Prouvost.
\newblock Machine learning for combinatorial optimization: a methodological
  tour d'horizon.
\newblock {\em arXiv preprint arXiv:1811.06128}, 2018.

\bibitem{cappart2019improving}
Quentin Cappart, Emmanuel Goutierre, David Bergman, and Louis-Martin Rousseau.
\newblock Improving optimization bounds using machine learning: Decision
  diagrams meet deep reinforcement learning.
\newblock In {\em Proceedings of the AAAI Conference on Artificial
  Intelligence}, volume~33, pages 1443--1451, 2019.

\bibitem{cappart2020combining}
Quentin Cappart, Thierry Moisan, Louis-Martin Rousseau, Isabeau
  Pr{\'e}mont-Schwarz, and Andre Cire.
\newblock Combining reinforcement learning and constraint programming for
  combinatorial optimization.
\newblock {\em arXiv preprint arXiv:2006.01610}, 2020.

\bibitem{cazenave2020polygames}
Tristan Cazenave, Yen-Chi Chen, Guan-Wei Chen, Shi-Yu Chen, Xian-Dong Chiu,
  Julien Dehos, Maria Elsa, Qucheng Gong, Hengyuan Hu, Vasil Khalidov, et~al.
\newblock Polygames: Improved zero learning.
\newblock {\em arXiv preprint arXiv:2001.09832}, 2020.

\bibitem{chen2019learning}
Xinyun Chen and Yuandong Tian.
\newblock Learning to perform local rewriting for combinatorial optimization.
\newblock In {\em Advances in Neural Information Processing Systems}, pages
  6278--6289, 2019.

\bibitem{christofides1976worst}
Nicos Christofides.
\newblock Worst-case analysis of a new heuristic for the travelling salesman
  problem.
\newblock Technical report, Carnegie-Mellon Univ Pittsburgh Pa Management
  Sciences Research Group, 1976.

\bibitem{cormen1990introduction}
Thomas~H Cormen, Charles~E Leiserson, Ronald~L Rivest, and Clifford Stein.
\newblock {\em Introduction to {A}lgorithms}.
\newblock MIT Press, 1990.

\bibitem{hanjun2017learning}
Hanjun Dai, Elias Khalil, Yuyu Zhang, Bistra Dilkina, and Le~Song.
\newblock Learning combinatorial optimization algorithms over graphs.
\newblock In {\em Advances in Neural Information Processing Systems}, pages
  6348--6358, 2017.

\bibitem{deudon2018learning}
Michel Deudon, Pierre Cournut, Alexandre Lacoste, Yossiri Adulyasak, and
  Louis-Martin Rousseau.
\newblock Learning heuristics for the {TSP} by policy gradient.
\newblock In {\em International Conference on the Integration of Constraint
  Programming, Artificial Intelligence, and Operations Research}, pages
  170--181. Springer, 2018.

\bibitem{dijkstra1959note}
Edsger~W Dijkstra.
\newblock A note on two problems in connexion with graphs.
\newblock {\em Numerische Mathematik}, 1(1):269--271, 1959.

\bibitem{drori2018alphad3m}
Iddo Drori, Yamuna Krishnamurthy, Remi Rampin, Raoni Lourenco, Jorge One,
  Kyunghyun Cho, Claudio Silva, and Juliana Freire.
\newblock Alpha{D3M}: {M}achine learning pipeline synthesis.
\newblock In {\em ICML International Workshop on Automated Machine Learning},
  2018.

\bibitem{emami2018learning}
Patrick Emami and Sanjay Ranka.
\newblock Learning permutations with sinkhorn policy gradient.
\newblock {\em arXiv preprint arXiv:1805.07010}, 2018.

\bibitem{erdHos1960evolution}
Paul Erd{\H{o}}s and Alfr{\'e}d R{\'e}nyi.
\newblock On the evolution of random graphs.
\newblock {\em Publ. Math. Inst. Hung. Acad. Sci}, 5(1):17--60, 1960.

\bibitem{gao2020learn}
Lei Gao, Mingxiang Chen, Qichang Chen, Ganzhong Luo, Nuoyi Zhu, and Zhixin Liu.
\newblock Learn to design the heuristics for vehicle routing problem.
\newblock {\em arXiv preprint arXiv:2002.08539}, 2020.

\bibitem{golden1980approximate}
Bruce Golden, Lawrence Bodin, T~Doyle, and W~Stewart~Jr.
\newblock Approximate traveling salesman algorithms.
\newblock {\em Operations Research}, 28(3-part-II):694--711, 1980.

\bibitem{ha2018world}
David Ha and J{\"u}rgen Schmidhuber.
\newblock World models.
\newblock {\em Conference on Neural Information Processing Systems}, 2018.

\bibitem{HOLLAND1983SBM}
Paul~W. Holland, Kathryn~Blackmond Laskey, and Samuel Leinhardt.
\newblock Stochastic blockmodels: First steps.
\newblock {\em Social Networks}, 5(2):109 -- 137, 1983.

\bibitem{hu2017solving}
Haoyuan Hu, Xiaodong Zhang, Xiaowei Yan, Longfei Wang, and Yinghui Xu.
\newblock Solving a new 3{D} bin packing problem with deep reinforcement
  learning method.
\newblock {\em arXiv preprint arXiv:1708.05930}, 2017.

\bibitem{huang2019coloring}
Jiayi Huang, Mostofa Patwary, and Gregory Diamos.
\newblock Coloring big graphs with alphagozero.
\newblock {\em arXiv preprint arXiv:1902.10162}, 2019.

\bibitem{joshi2019efficient}
Chaitanya~K Joshi, Thomas Laurent, and Xavier Bresson.
\newblock An efficient graph convolutional network technique for the travelling
  salesman problem.
\newblock {\em arXiv preprint arXiv:1906.01227}, 2019.

\bibitem{kahneman2011thinking}
Daniel Kahneman.
\newblock {\em Thinking, fast and slow}.
\newblock Macmillan, 2011.

\bibitem{karger1995randomized}
David~R Karger, Philip~N Klein, and Robert~E Tarjan.
\newblock A randomized linear-time algorithm to find minimum spanning trees.
\newblock {\em Journal of the ACM (JACM)}, 42(2):321--328, 1995.

\bibitem{kim2003generating}
Jeong~Han Kim and Van~H Vu.
\newblock Generating random regular graphs.
\newblock In {\em Proceedings of the thirty-fifth annual ACM symposium on
  Theory of computing}, pages 213--222, 2003.

\bibitem{king1997simpler}
Valerie King.
\newblock A simpler minimum spanning tree verification algorithm.
\newblock {\em Algorithmica}, 18(2):263--270, 1997.

\bibitem{kool2019attention}
Wouter Kool, Herke van Hoof, and Max Welling.
\newblock Attention, learn to solve routing problems!
\newblock {\em International Conference on Learning Representations}, 2019.

\bibitem{kruskal1956shortest}
Joseph~B Kruskal.
\newblock On the shortest spanning subtree of a graph and the traveling
  salesman problem.
\newblock {\em Proceedings of the American Mathematical Society}, 7(1):48--50,
  1956.

\bibitem{laterre2018ranked}
Alexandre Laterre, Yunguan Fu, Mohamed~Khalil Jabri, Alain-Sam Cohen, David
  Kas, Karl Hajjar, Torbjorn~S Dahl, Amine Kerkeni, and Karim Beguir.
\newblock Ranked reward: Enabling self-play reinforcement learning for
  combinatorial optimization.
\newblock {\em arXiv preprint arXiv:1807.01672}, 2018.

\bibitem{li2018combinatorial}
Zhuwen Li, Qifeng Chen, and Vladlen Koltun.
\newblock Combinatorial optimization with graph convolutional networks and
  guided tree search.
\newblock In {\em Advances in Neural Information Processing Systems}, pages
  539--548, 2018.

\bibitem{liao2020deep}
Haiguang Liao, Wentai Zhang, Xuliang Dong, Barnabas Poczos, Kenji Shimada, and
  Levent Burak~Kara.
\newblock A deep reinforcement learning approach for global routing.
\newblock {\em Journal of Mechanical Design}, 142(6), 2020.

\bibitem{lin1965computer}
Shen Lin.
\newblock Computer solutions of the traveling salesman problem.
\newblock {\em Bell System Technical Journal}, 44(10):2245--2269, 1965.

\bibitem{lin1973effective}
Shen Lin and Brian~W Kernighan.
\newblock An effective heuristic algorithm for the traveling-salesman problem.
\newblock {\em Operations Research}, 21(2):498--516, 1973.

\bibitem{lu2020vrp}
Hao Lu, Xingwen Zhang, and Shuang Yang.
\newblock A learning-based iterative method for solving vehicle routing
  problems.
\newblock In {\em International Conference on Learning Representations}, 2020.

\bibitem{ma2020combinatorial}
Qiang Ma, Suwen Ge, Danyang He, Darshan Thaker, and Iddo Drori.
\newblock Combinatorial optimization by graph pointer networks and hierarchical
  reinforcement learning.
\newblock {\em AAAI Workshop on Deep Learning on Graphs: Methodologies and
  Applications}, 2020.

\bibitem{malazgirt2019tauriel}
Gorker~Alp Malazgirt, Osman~S Unsal, and Adrian~Cristal Kestelman.
\newblock Tau{R}iel: Targeting traveling salesman problem with a deep
  reinforcement learning inspired architecture.
\newblock {\em arXiv preprint arXiv:1905.05567}, 2019.

\bibitem{mazyavkina2020reinforcement}
Nina Mazyavkina, Sergey Sviridov, Sergei Ivanov, and Evgeny Burnaev.
\newblock Reinforcement learning for combinatorial optimization: A survey.
\newblock {\em arXiv preprint arXiv:2003.03600}, 2020.

\bibitem{nazari2018reinforcement}
Mohammadreza Nazari, Afshin Oroojlooy, Lawrence Snyder, and Martin Tak{\'a}c.
\newblock Reinforcement learning for solving the vehicle routing problem.
\newblock In {\em Advances in Neural Information Processing Systems}, pages
  9839--9849, 2018.

\bibitem{nevsetvril2001otakar}
Jaroslav Nešetřil, Eva Milková, and Helena Nešetřilová.
\newblock Otakar {B}orůvka on minimum spanning tree problem translation of
  both the 1926 papers, comments, history.
\newblock {\em Discrete Mathematics}, 233(1):3 -- 36, 2001.
\newblock Czech and Slovak 2.

\bibitem{rltaxanomy2020openai}
{OpenAI}.
\newblock {Spinning Up in Deep RL}.
\newblock
  \url{https://spinningup.openai.com/en/latest/spinningup/rl_intro2.html},
  2020.

\bibitem{gurobi2019}
Gurobi {O}ptimization.
\newblock Gurobi solver.
\newblock \url{https://www.gurobi.com/pdfs/benchmarks.pdf}, 2019.

\bibitem{parker2020alphatsp}
Felix Parker and Darius Irani.
\newblock {A}lpha{TSP}: Learning a {TSP} {H}euristic using the {A}lpha{Z}ero
  methodology.
\newblock {\em JHU {T}echnical {R}eport}, 2020.

\bibitem{pierrot2019learning}
Thomas Pierrot, Guillaume Ligner, Scott~E Reed, Olivier Sigaud, Nicolas Perrin,
  Alexandre Laterre, David Kas, Karim Beguir, and Nando de~Freitas.
\newblock Learning compositional neural programs with recursive tree search and
  planning.
\newblock In {\em Advances in Neural Information Processing Systems}, pages
  14646--14656, 2019.

\bibitem{prim1957shortest}
Robert~Clay Prim.
\newblock Shortest connection networks and some generalizations.
\newblock {\em The Bell System Technical Journal}, 36(6):1389--1401, 1957.

\bibitem{tsplib}
Gerhard Reinelt.
\newblock {TSPLIB: {L}ibrary of sample instances for the TSP}, 2020.

\bibitem{rosenkrantz1977analysis}
Daniel~J Rosenkrantz, Richard~E Stearns, and Philip~M Lewis, II.
\newblock An analysis of several heuristics for the traveling salesman problem.
\newblock {\em {SIAM} {J}ournal on {C}omputing}, 6(3):563--581, 1977.

\bibitem{schrittwieser2019mastering}
Julian Schrittwieser, Ioannis Antonoglou, Thomas Hubert, Karen Simonyan,
  Laurent Sifre, Simon Schmitt, Arthur Guez, Edward Lockhart, Demis Hassabis,
  Thore Graepel, et~al.
\newblock Mastering {A}tari, {G}o, chess and {S}hogi by planning with a learned
  model.
\newblock {\em arXiv preprint arXiv:1911.08265}, 2019.

\bibitem{selsam2018learning}
Daniel Selsam, Matthew Lamm, Benedikt B{\"u}nz, Percy Liang, Leonardo de~Moura,
  and David~L Dill.
\newblock Learning a {SAT} solver from single-bit supervision.
\newblock {\em International Conference on Learning Representations}, 2019.

\bibitem{silver2017mastering}
David Silver, Julian Schrittwieser, Karen Simonyan, Ioannis Antonoglou, Aja
  Huang, Arthur Guez, Thomas Hubert, Lucas Baker, Matthew Lai, Adrian Bolton,
  et~al.
\newblock Mastering the game of {Go} without human knowledge.
\newblock {\em Nature}, 550(7676):354, 2017.

\bibitem{song2019co}
Jialin Song, Ravi Lanka, Yisong Yue, and Masahiro Ono.
\newblock Co-training for policy learning.
\newblock {\em Uncertainty in {AI}}, 2019.

\bibitem{steger1999generating}
Angelika Steger and Nicholas~C Wormald.
\newblock Generating random regular graphs quickly.
\newblock {\em Combinatorics, Probability and Computing}, 8(4):377--396, 1999.

\bibitem{sun2018dual}
Wen Sun, Geoffrey~J Gordon, Byron Boots, and J~Bagnell.
\newblock Dual policy iteration.
\newblock In {\em Advances in Neural Information Processing Systems}, pages
  7059--7069, 2018.

\bibitem{tang2019reinforcement}
Yunhao Tang, Shipra Agrawal, and Yuri Faenza.
\newblock Reinforcement {L}earning for {I}nteger {P}rogramming: {L}earning to
  {C}ut.
\newblock {\em arXiv preprint arXiv:1906.04859}, 2019.

\bibitem{velivckovic2017graph}
Petar Veli{\v{c}}kovi{\'c}, Guillem Cucurull, Arantxa Casanova, Adriana Romero,
  Pietro Lio, and Yoshua Bengio.
\newblock Graph attention networks.
\newblock {\em International Conference on Learning Representations}, 2018.

\bibitem{velikovi2020neural}
Petar Veličković, Rex Ying, Matilde Padovano, Raia Hadsell, and Charles
  Blundell.
\newblock Neural execution of graph algorithms.
\newblock In {\em International Conference on Learning Representations}, 2020.

\bibitem{vesselinova2020learning}
Natalia Vesselinova, Rebecca Steinert, Daniel~F. Perez-Ramirez, and Magnus
  Boman.
\newblock Learning combinatorial optimization on graphs: {A} survey with
  applications to networking.
\newblock {\em arXiv preprint arXiv:2005.11081}, 2020.

\bibitem{vinyals2015pointer}
Oriol Vinyals, Meire Fortunato, and Navdeep Jaitly.
\newblock Pointer networks.
\newblock In {\em Advances in {N}eural {I}nformation {P}rocessing {S}ystems},
  pages 2692--2700, 2015.

\bibitem{watts1998collective}
Duncan~J Watts and Steven~H Strogatz.
\newblock Collective {D}ynamics of ‘{S}mall-{W}orld’ {N}etworks.
\newblock {\em {N}ature}, 393(6684):440, 1998.

\bibitem{williamson2011design}
David~P Williamson and David~B Shmoys.
\newblock {\em The {D}esign of {A}pproximation {A}lgorithms}.
\newblock Cambridge {U}niversity {P}ress, 2011.

\bibitem{xu2018powerful}
Keyulu Xu, Weihua Hu, Jure Leskovec, and Stefanie Jegelka.
\newblock How powerful are graph neural networks?
\newblock {\em International Conference on Learning Representations}, 2019.

\bibitem{xu2020reason}
Keyulu Xu, Jingling Li, Mozhi Zhang, Simon~S. Du, Ken-ichi Kawarabayashi, and
  Stefanie Jegelka.
\newblock What can neural networks reason about?
\newblock In {\em International Conference on Learning Representations}, 2020.

\bibitem{xu2019learning}
Ruiyang Xu and Karl Lieberherr.
\newblock Learning self-game-play agents for combinatorial optimization
  problems.
\newblock In {\em {P}roceedings of the 18th {I}nternational {C}onference on
  {A}utonomous {A}gents and {M}ulti{A}gent {S}ystems}, pages 2276--2278, 2019.

\bibitem{yolcu2019learning}
Emre Yolcu and Barnabas Poczos.
\newblock {L}earning local search heuristics for boolean satisfiability.
\newblock In {\em Advances in Neural Information Processing Systems}, pages
  7990--8001, 2019.

\end{thebibliography}


\end{document}